\documentclass[sigconf,nonacm]{acmart}
\AtBeginDocument{%
  }

\usepackage{algorithm}
\usepackage{algpseudocode}

\usepackage{xspace}
\usepackage{multirow}
\usepackage{xcolor}
\usepackage{subcaption}
\newcommand{\base}{\textsc{Base}\xspace}
\newcommand{\basex}[1]{\textsc{Base+#1}\xspace}
\newcommand{\lantern}{\textsc{Lantern}\xspace}

\begin{document}

\title{Lantern: Conflict-Aware Gradient Blending for Physics-Guided Diffusion Models in Calorimeter Simulation}

\author{Farzana Yasmin Ahmad}
\email{fa7sa@virginia.edu}
\correspondingauthor
\affiliation{%
  \institution{Computer Science Department, University of Virginia}
  \country{USA}
}
\author{Vanamala Venkataswamy}
\email{vv3xu@virginia.edu}
\affiliation{%
  \institution{Visiting Scholar, University of Virginia}
  \country{USA}
}

\author{Geoffrey Fox}
\email{gcfexchange@gmail.com}
\affiliation{%
  \institution{Computer Science Department, Biocomplexity Institute, University of Virginia}
  \country{USA}
}

\renewcommand{\shortauthors}{Ahmad et al.}

\begin{abstract}
Monte Carlo simulation of calorimeter showers is a principal computational bottleneck for the High-Luminosity LHC, and diffusion models have emerged as fast, high-fidelity surrogates. Their denoising objective is purely statistical, however: a model can minimize it while placing the physics wrong. Existing physics-informed generative methods cannot close this gap, because they assume a closed-form law, a governing PDE residual or a hard per-sample constraint, that a shower does not supply: no per-sample PDE governs a stochastic cascade, and energy conservation fixes only one scalar per shower. Standard metrics share a blind spot of their own, comparing showers in a physics feature space while ignoring the correlation structure across calorimeter layers and voxels. We address both gaps. We introduce the Correlation Frobenius Distance (CFD), a single normalized score for correlation fidelity at layer-wise and voxel-wise scales. We then encode the soft per-sample structure available in a shower as two physics-aware auxiliary losses. The first is a variance-stabilized voxel residual loss grounded in counting statistics; the second is a graph Laplacian loss over the detector geometry. We combine both with denoising through GradBlend, which anchors the step magnitude to the denoising gradient while letting the auxiliary steer its direction, yielding \lantern{}, a physics-guided diffusion surrogate. On Dataset~2 of the CaloChallenge, injecting the physics losses through task-symmetric rules such as PCGrad, GradNorm, IMTL-G, and ConFIG inflates FPD by $2$--$100\times$ relative to denoising alone, whereas GradBlend admits the same signal without regression and, with the Laplacian loss, \lantern{} improves both FPD and CFD. Our ablation study on the auxiliary loss scheduler shows that the voxel residual loss, whose gradient conflicts with denoising, requires a terminal denoising-only phase to preserve shower fidelity, whereas the non-conflicting Laplacian loss is insensitive to the schedule.
\end{abstract}



\maketitle


\section{Introduction}
\label{sec:intro}

Calorimeter shower simulation models the particle cascade produced when
a high-energy particle strikes the detector material, and underpins
detector design, calibration, and event reconstruction in high-energy
physics. The standard tools are Monte Carlo simulators such as
Geant4~\cite{agostinelli2003geant4}, accurate but computationally
expensive; the upcoming High-Luminosity upgrade of the Large Hadron
Collider~\cite{apollinari2017high} makes Geant4 alone impractical at the
required data volume. The CaloChallenge benchmark~\cite{calochallenge_homepage}
spurred deep generative surrogates for full simulation, many producing
high-fidelity showers at a fraction of the cost.

Evaluating these models is nontrivial: a faithful surrogate must
preserve the underlying physics, not only visual plausibility. The
standard distributional metrics, the Fréchet and kernel physics
distances (FPD and KPD)~\cite{kansal2023evaluating}, compare samples in
a physics feature space and detect global mismatch, but do not test
whether a model reproduces correlations between calorimeter layers and
voxels; the Pearson Correlation Coefficient (PCC)~\cite{ahmad2024comprehensive}
probes these correlations but returns a full matrix rather than a single
comparable score. We introduce the Correlation Frobenius Distance (CFD), a single
normalized score for layer- and voxel-wise correlation fidelity.

CFD diagnoses where a surrogate departs from Geant4 after training; a
complementary question is whether physics structure can be built into
training itself. The denoising objective of diffusion is purely
statistical, and existing physics-informed variants assume a closed-form
law that a stochastic shower does not supply (Section~\ref{sec:related}).
Physics-guided training has consequently entered shower generation only
through coarse aggregates: the concurrent
CaloTrilogy~\cite{jiang2026calotrilogy} penalizes layer-wise energies
with a scalar-weighted sum and offers no mechanism to keep the physics
term from degrading the generative objective. We address both gaps: voxel-level losses and a combination rule that protects denoising.

Although no conservation law governs a shower voxel by voxel, each
sample still carries physical structure: voxel energies whose
fluctuations scale with the deposited energy, and local differences
between neighboring cells set by the detector geometry. We encode these
as two physics-aware auxiliary losses: a variance-stabilized voxel
residual loss and a graph Laplacian loss over face-sharing voxels
(Section~\ref{sec:method}). Pairing either loss with denoising turns
training into a multi-task problem, and existing combination rules treat
the objectives as peers, none protecting denoising when an auxiliary
pulls against it. Since denoising keeps the model generative, our second
contribution is a gradient blending rule that admits the physics
gradient while keeping denoising primary. We study these questions on
Dataset 2 of the CaloChallenge~\cite{calochallenge_homepage}, a
high-granularity calorimeter of $6480$ voxels.

\paragraph{Contributions.}
This paper delivers a metric, two losses, and a combination rule that
together make physics guidance safe for a strong generative surrogate.
(1)~We introduce the \emph{Correlation Frobenius Distance} (CFD), to our knowledge, the first single-score metric for shower correlation structure that FPD and KPD miss, at layer-wise and voxel-wise scales. (2)~We design two physics-aware auxiliary losses, a variance-stabilized voxel residual loss and a graph Laplacian loss, encoding soft per-sample structure that no closed-form constraint expresses. (3)~We introduce \textsc{GradBlend}, a denoising-anchored blending rule: the auxiliary steers only the update direction while the magnitude stays anchored to denoising, keeping the generative objective primary by construction. (4)~We show this anchoring is necessary: PCGrad, GradNorm, IMTL-G, and ConFIG degrade FPD by $2$--$100\times$, while GradBlend admits the physics signal and, with the Graph Laplacian loss, improves both FPD and CFD. (5)~Because the auxiliary weight cancels from GradBlend's update, the schedule reduces to a simple on/off gate; we show its terminal denoising-only phase is essential for sample coverage.


\section{Related Work}
\label{sec:related}


Diffusion-based surrogates learn a data distribution through a
single denoising objective, equivalent to learning its
score~\cite{ho2020denoising, song2021scorebased}. The generality that
lets diffusion fit any distribution from samples alone leaves the
physics unconstrained: a surrogate can minimize the loss while
misplacing the relationships that make its outputs physically valid.
This has motivated writing physics directly into generative training,
through strategies that calorimeter showers fit none of. 
PDE-residual losses~\cite{bastek2025pidm, baldan2025flow, zhang2025physics},
inference-time guidance and projection~\cite{chung2023dps, gao2023prediff,
liang2025ccfm, trupin2026learning, utkarsh2026physics}, equivariant
architectures~\cite{hassan2024etflow, hoogeboom2022edm, tian2025equiflow},
and energy-based steering~\cite{li2026elign, vaitl2025pathgrad} each
presuppose a closed-form per-sample object: a governing equation,
differentiable constraint, symmetry group, or tractable energy.
A calorimeter shower
supplies none. It is a single stochastic Geant4 realization whose
physical content lives in the ensemble, and its only hard per-sample
constraints, total and per-layer energy, are already imposed by
construction through conditioning and preprocessing
(Section~\ref{sec:formulation}), leaving no per-sample target to
enforce. Even where a physics loss applies, adding it to a generative
objective is fragile: naive weighted sums produce pathological
training dynamics when gradient scales and directions
disagree~\cite{wang2021pathologies, liu2025config}. How the physics
gradient combines with the generative gradient is therefore not an
implementation detail but the central obstacle.

Expressing ensemble structure as an auxiliary loss alongside denoising
turns training into a two-objective gradient problem, for which
multi-task learning supplies combination rules.
PCGrad~\cite{yu2020pcgrad}, GradNorm~\cite{chen2018gradnorm},
IMTL-G~\cite{liu2021towards}, and ConFIG~\cite{liu2025config} treat the
objectives as peers, granting each equal influence over update
direction or training rate. Peer treatment suits co-equal tasks, not a
setting where one objective must remain primary: splitting the
direction evenly forces denoising to cede half of each update to the
auxiliary, suppressing the objective that makes the model generative.
A separate line treats one task as primary: gradient cosine similarity (GCS) gates the auxiliary loss by the sign or value of its cosine with the main-task gradient, with guaranteed convergence to critical points of the main task~\cite{du2018gcs}, and this primary-anchored gating has since shown gains in physics-informed PDE solving~\cite{yan2023auxiliary}. GradBlend belongs to this
primary-anchored family but differs in mechanism: GCS and its
successors modulate the auxiliary's \emph{scalar weight} while summing
raw gradients, so a large-norm auxiliary still dominates the update
direction. GradBlend instead separates direction from magnitude: it
blends unit directions so neither gradient wins by norm, anchors the
step size to the denoising gradient alone, and gates the auxiliary's
directional share by the conflict angle.

Within calorimeter simulation, surrogates almost universally optimize a
data-driven objective, encoding physics through architecture rather
than the training loss~\cite{amram2023calodiffusion, birk2024omnijet,
buhmann2024caloclouds2, favaro2025calodream, kobylianskii2024calograph,
leigh2024pcjedi, leigh2024pcdroid, mikuni2022score,
mikuni2024caloscore2}. The one exception,
CaloTrilogy~\cite{jiang2026calotrilogy}, adds a per-layer energy
constraint to a MeanFlow generator, weighted by a fixed coefficient or
an adaptive Lagrange multiplier from the modified differential method
of multipliers~\cite{platt1988mdmm}. Its constraint has a per-sample
target (each layer energy has a true value), so once the representation
stabilizes the physics and generative gradients agree and scalar
magnitude control suffices; still, its reported sensitivity to
constraint scheduling shows the combination is delicate in the
weakly-conflicting regime. The auxiliaries we study have no per-sample
target: they encode correlation structure that per-sample marginals
cannot express, whose optimum opposes denoising and places the problem
in the conflicting regime. To our knowledge, no prior calorimeter
surrogate has treated physics-auxiliary training as a
gradient-combination problem; characterizing that conflict and
resolving it with a denoising-anchored rule is our contribution.


\section{Problem Formulation}
\label{sec:formulation}

We represent a calorimeter shower in physical
units (MeV) as $x \in \mathbb{R}_{\ge 0}^{V}$ where $V$ is the number of voxels in a shower and its preprocessed form as
$\tilde{x} = \mathcal{T}(x) \in \mathbb{R}^{V}$, where $\mathcal{T}$ is a composition of transformation steps. The network works in the preprocessed space $\tilde{x}$,
and the inverse transform $\mathcal{T}^{-1}$ maps a sample back to physical
energy $x$ when a physical quantity is needed. Any quantity with a time index
or a hat ($x_t$, $\hat{x}_0$) is in the preprocessed space; the bare symbol
$x$ means physical energy.

We treat simulation as conditional generation. \textsc{Geant4} defines the
target distribution $p(x \mid c)$, where $x$ is the shower in physical space and
$c$ is the conditioning information. Following prior work~\cite{mikuni2024caloscore2,favaro2025calodream}, we factorize the generation into two stages, each conditioned on its
own subset of $c$, and train a separate diffusion model for each. The energy
network, with parameters $\phi$, models $p_\phi(u \mid E_\text{inc})$, where
$E_\text{inc}$ is the incident energy of the shower and $u \in \mathbb{R}^{L}$ is
derived from the per-layer energies of the shower across its $L$ longitudinal
layers. The shape network, with parameters $\theta$, models
$p_\theta(x \mid E_\text{inc}, u)$, the voxelized shower given the incident
energy and $u$. Marginalizing over $u$ recovers the target:
\begin{equation}
p(x \mid E_\text{inc}) = \int p_\theta(x \mid E_\text{inc}, u)\,
  p_\phi(u \mid E_\text{inc})\, \mathrm{d}u .
\end{equation}
The two networks are trained independently and composed at generation time. Our
auxiliary losses and gradient method act on the shape network alone; the energy
network keeps the standard denoising objective. During shape-network training,
$u$ is the ground-truth value, so the network conditions on exact layer ratios,
while at generation $u$ is drawn from the energy network.

\paragraph{Denoising objectives.}
Both networks are denoising diffusion probabilistic models
(DDPM)~\cite{ho2020denoising}. The forward process perturbs a clean sample $z_0$
under a variance schedule $\beta_1,\dots,\beta_T$; writing
$\alpha_t = 1-\beta_t$ and
$\bar{\alpha}_t = \prod_{s=1}^{t}\alpha_s$, the noised state at
timestep $t \in \{1,\dots,T\}$ is
\begin{equation}
  z_t = \sqrt{\bar{\alpha}_t}\, z_0
        + \sqrt{1-\bar{\alpha}_t}\, \epsilon,
  \qquad \epsilon \sim \mathcal{N}(0, I).
  \label{eq:forward}
\end{equation}
The energy network applies Eq.~\eqref{eq:forward} to the clean
ratio vector $u$ and minimizes the standard denoising loss over
$\epsilon_\phi(u_t, t, E_{\mathrm{inc}})$; it generates the ratios
autoregressively following CaloDREAM~\cite{favaro2025calodream}, carries no auxiliary
loss, and is detailed in Appendix~\ref{app:energy}. All contributions of this
paper act on the shape network, which applies
Eq.~\eqref{eq:forward} to the preprocessed shower $\tilde{x}_0$
and minimizes, with conditioning $(E_{\mathrm{inc}}, u)$,
\begin{equation}
  \mathcal{L}_{\mathrm{denoise}}(\theta)
  = \mathbb{E}_{\tilde{x}_0,\epsilon,t}
    \left[ \left\lVert \epsilon
      - \epsilon_\theta(x_t, t, E_{\mathrm{inc}}, u)
      \right\rVert_2^2 \right].
  \label{eq:denoise-loss}
\end{equation}

\paragraph{Physics-aware auxiliary objective.}
The denoising loss scores the predicted noise, not the shower. To reason about
the shower itself, we use the model's single-step clean estimate
\begin{equation}
\hat{x}_0(x_t, t) =
\frac{x_t - \sqrt{1 - \bar{\alpha}_t}\,\epsilon_\theta(x_t, t, E_\text{inc}, u)}
     {\sqrt{\bar{\alpha}_t}}.
\label{eq:x0hat}
\end{equation}
We add a physics-aware auxiliary loss on this estimate,
\begin{equation}
\mathcal{L}_\text{aux}(\theta) =
\mathbb{E}_{\tilde{x}_0,\,\epsilon,\,t}\!\left[\,
\ell_\text{phys}\big(\hat{x}_0(x_t, t),\, \tilde{x}_0\big)
\,\right],
\label{eq:aux}
\end{equation}
where $\ell_\text{phys}$ is a per-sample penalty that compares the clean estimate
with the true shower. We instantiate $\ell_\text{phys}$ and describe how it is
combined with the denoising objective in Section~\ref{sec:method}.


The denoising and auxiliary objectives share one update on the shape network.
Writing their gradients as
\begin{equation}
g_d = \nabla_\theta \mathcal{L}_\text{denoise},\,\,\,
g_a = \nabla_\theta \big( w_\text{aux}(s)\,\mathcal{L}_\text{aux} \big)
    = w_\text{aux}(s)\, \nabla_\theta \mathcal{L}_\text{aux},
\label{eq:gradients}
\end{equation}
where $w_\text{aux}(s) \ge 0$ weights the auxiliary loss at optimization
step $s$ and the denoising loss is never reweighted, we obtain a
multi-task gradient problem in which $g_d$ and $g_a$ need not agree.
Denoising is the objective that makes the network generative, so it must
stay primary; an auxiliary that overrides it degrades the samples it is
meant to improve. The problem is therefore twofold. First, construct an
update $g_\text{blend}(g_d, g_a)$ that admits the physics signal while
keeping denoising primary. Second, decide when during training the
auxiliary should act. We address the first with GradBlend
(Section~\ref{sec:gradblend}) and the second with the schedule
$w_\text{aux}(s)$ (Section~\ref{sec:schedule}).
\section{Method}
\label{sec:method}

\label{sec:architecture}
Our surrogate follows the factorized design of
CaloDREAM~\cite{favaro2025calodream}, with the energy and shape networks
of Section~\ref{sec:formulation}; unlike CaloDREAM's conditional flow
matching, we use DDPM~\cite{ho2020denoising} for both. The energy
network is an autoregressive transformer over the $L$ layers: the ratio
$u_i$ for layer $i$ is predicted from $E_{\mathrm{inc}}$ and the
preceding layers through causal masking, matching the forward-propagating
structure of a shower. The shape network, which carries all of our
contributions, is a conditional Vision
Transformer~\cite{dosovitskiy2020image, peebles2023scalable} trained to
predict the injected noise $\epsilon_\theta(x_t, t, E_{\mathrm{inc}}, u)$:
it patch-embeds the calorimeter volume, processes the patches with
transformer blocks conditioned on $t$, $E_{\mathrm{inc}}$, and $u$ through
affine modulation, and projects back to voxel space. The preprocessing
transform $\mathcal{T}$ (Eq.~\eqref{eq:preproc}) is taken unchanged from
CaloDREAM; full architectural details and hyperparameters are in
Appendix~\ref{app:arch}.



\subsection{Physics-Aware Auxiliary Losses}
\label{sec:aux-losses}

\subsubsection{Variance-Stabilized Voxel Residual Loss}
\label{sec:voxel-loss}

Here we take $\ell_\text{phys}$ as a per-voxel residual loss. The denoising loss Eq.~\eqref{eq:denoise-loss} weights every voxel equally, but not all calorimeter voxels are equally informative. In a calorimeter shower a few bright cells carry most of the energy, while many dim cells fix the sparsity pattern and the low-energy tail.
 Shower signals are counting processes, so the fluctuation scale of a voxel grows as the square root of its physical energy \cite{fabjan2003calorimetry}. This motivates an inverse-variance weighting that standardizes each residual by an energy-dependent scale, moving the capacity from the bright core toward the dim tail.
\begin{equation}
  r_i = \frac{\hat{x}_{0,i} - \tilde{x}_{0,i}}
             {\sqrt{\lvert \tilde{x}_{0,i} \rvert + \epsilon_{\mathrm{vox}}}},
  \qquad
  \ell_{\mathrm{vox}}
  = \frac{1}{V} \sum_{i=1}^{V}
    \mathrm{Huber}_{\delta_{\mathrm{vox}}}(r_i),
  \label{eq:voxel-loss}
\end{equation}
where $i \in \{1, \dots, V\}$ indexes voxels and $\hat{x}_{0,i}$ and
$\tilde{x}_{0,i}$ are the $i$-th voxel of the model's clean estimate
and of the Geant4 reference, with floor
$\epsilon_{\mathrm{vox}} = 10^{-6}$ and Huber transition
$\delta_{\mathrm{vox}} = 1$. The true values $\tilde{x}_{0,i}$ are
detached from the computational graph, so the denominator acts as a
constant per-voxel weight rather than an optimization target. The
Huber penalty
\begin{equation}
  \mathrm{Huber}_\delta(r) =
  \begin{cases}
    \tfrac{1}{2} r^2 & \lvert r \rvert \le \delta, \\
    \delta \left( \lvert r \rvert - \tfrac{1}{2}\delta \right)
      & \lvert r \rvert > \delta,
  \end{cases}
  \label{eq:huber}
\end{equation}
stays quadratic on the bulk of residuals and linear beyond
$\delta_{\mathrm{vox}}$, so voxels whose residuals exceed the assumed
scale bound their influence on the update rather than dominating it.

\subsubsection{Graph-Laplacian Loss}
\label{sec:laplacian-loss}
The variance-stabilized voxel residual loss of
Section~\ref{sec:voxel-loss} is pointwise: it ignores how each voxel
relates to its neighbors, so a model can match every voxel on average
yet still get the local correlations wrong. The second auxiliary loss
compares the predicted and reference neighborhood structure across
physically adjacent voxels.

We represent the calorimeter as a graph $G$ with one node per voxel,
connected to face-sharing neighbors along the three grid axes: periodic
in azimuth $\varphi$, open radially and longitudinally. Interior voxels
have six neighbors; boundary voxels have fewer, except along $\varphi$,
which always supplies both. The combinatorial Laplacian $L = D - A$
(adjacency $A$, degree $D$), applied to a voxel signal $y$, returns at
each node the summed difference to its neighbors,
$(Ly)_i = \sum_{j \sim i} (y_i - y_j)$, a measure of local roughness.

In graph signal processing the Laplacian acts as a high-pass
filter~\cite{shuman2013emerging}: it responds weakly where neighboring
voxels agree and strongly where they differ, picking out edges rather
than flat regions. A standard regularizer minimizes the Dirichlet energy
$\hat{x}_0^\top L \hat{x}_0$ to drive the signal toward smoothness, but a
calorimeter shower is not smooth, so flattening it would erase the
structure. We therefore penalize the gap between the predicted and
reference Laplacian responses,
\begin{equation}
\ell_\text{lap} = \frac{1}{\lambda_\text{max}(L)^2}\,
  \frac{1}{V}\big\| L\hat{x}_0 - L\tilde{x}_0 \big\|_2^2 ,
\label{eq:laplacian-loss}
\end{equation}
which is zero when the model reproduces Geant4's local differences.

The prefactor $1/\lambda_{\max}(L)^2$ keeps the auxiliary gradient on a
scale comparable to the denoising gradient, as the combination scheme of
Section~\ref{sec:gradblend} assumes: the squared Laplacian norm inflates
the loss by up to $\lambda_{\max}(L)^2$, so dividing by
$\lambda_{\max}(L)^2 \approx 95.5$ ($\lambda_{\max}(L) \approx 9.77$,
computed once and cached) bounds the operator gain by one. $L$ is a fixed
sparse operator applied twice per step, adding only $O(V)$ cost and no
second forward pass.

\subsection{GradBlend: A Denoising-Anchored Combination}
\label{sec:gradblend}
Section~\ref{sec:formulation} posed the combination problem: build a
single update from the denoising gradient $g_d$ and the auxiliary
gradient $g_a$ without letting the auxiliary displace denoising. A
rule that treats the two gradients as peers fails this under
conflict. GradBlend separates the update into a \emph{direction} and
a \emph{magnitude} and assigns the two gradients different roles in
each: the auxiliary may steer the direction, but the magnitude is
anchored to denoising alone.

Let $\hat{g}_d = g_d / \lVert g_d \rVert$ and
$\hat{g}_a = g_a / \lVert g_a \rVert$ be the unit directions of the
two gradients, and let
\begin{equation}
  \theta_{\mathrm{conf}} = \arccos\!\left(\hat{g}_d \cdot \hat{g}_a\right)
  \label{eq:conflict-angle}
\end{equation}
be the \emph{conflict angle} between them: below $90^\circ$ the
objectives reinforce, above $90^\circ$ they oppose. A coefficient $\rho$, fixed to $1$ throughout, sets the auxiliary's share of a
tentative blend direction,
\begin{equation}
  \hat{g}_{\mathrm{tent}}
  = \frac{\hat{g}_d + \rho\, \hat{g}_a}
         {\lVert \hat{g}_d + \rho\, \hat{g}_a \rVert},
  \qquad
  \cos\varphi_{\mathrm{align}} = \hat{g}_d \cdot \hat{g}_{\mathrm{tent}},
  \label{eq:tentative-blend}
\end{equation}
where the \emph{alignment angle} $\varphi_{\mathrm{align}}$ measures how far the
tentative blend rotates away from the denoising direction. With $\rho = 1$ the
tentative blend is the bisector of the two unit directions, and the two angles
are related in closed form,
\begin{equation}
  \cos\varphi_{\mathrm{align}}
  = \sqrt{\frac{1 + \cos\theta_{\mathrm{conf}}}{2}}
  = \cos\!\left(\frac{\theta_{\mathrm{conf}}}{2}\right),
  \label{eq:closed-form}
\end{equation}
so the entire construction is parameterized by the single conflict
angle $\theta_{\mathrm{conf}}$.

\paragraph{Magnitude.}
The step magnitude is anchored to the denoising gradient and
contracts with conflict:
\begin{equation}
  m_{\mathrm{ref}}
  = \lVert g_d \rVert \cdot
    \max\!\left(\cos\varphi_{\mathrm{align}},\, 0.05\right)
  = \lVert g_d \rVert \cdot
    \max\!\left(\cos\tfrac{\theta_{\mathrm{conf}}}{2},\, 0.05\right).
  \label{eq:magnitude}
\end{equation}
The auxiliary enters only through the direction, never the step size:
the update equals a pure denoising step when the two gradients agree
and shrinks as they conflict, so the auxiliary cannot inflate the
effective learning rate, and the update stays conservative under
conflict. The floor of $0.05$ keeps the step from collapsing under
near-antiparallel gradients.

\paragraph{Direction.}
The auxiliary contributes to the direction only when the conflict is
moderate. Its effective coefficient is gated by
$\theta_{\mathrm{conf}}$: held at full weight below $120^\circ$,
decayed linearly to zero between $120^\circ$ and $150^\circ$, and
suppressed above. The final direction is the unit-normalized
$\rho_{\mathrm{eff}}$-weighted blend of $\hat{g}_d$ and $\hat{g}_a$;
because both inputs are unit vectors, neither gradient can dominate the
direction by raw norm.

Note that the magnitude rule~\eqref{eq:magnitude} uses the
\emph{ungated} alignment of Eq.~\eqref{eq:tentative-blend}, computed
at the full coefficient $\rho$, even in regimes where the gate
reduces or removes the auxiliary from the direction. This is
deliberate: a large measured conflict signals that the auxiliary
disagrees strongly with denoising in the current region of parameter
space, and we shrink the step as a stability precaution even when the
conflicting direction is discarded. It also decouples the magnitude
rule from the gate thresholds, keeping the step size a smooth
function of $\theta_{\mathrm{conf}}$ alone. Earlier variants that
coupled the magnitude to the auxiliary norm were unstable in
training; this rule stabilized them.

\paragraph{The GradBlend update.}
Algorithm~\ref{alg:gradblend} gives the full construction.\footnote{The
implementation adds one numerical guard: it also falls back to
$\hat{g}_d$ if the blended direction is nearly degenerate,
$\lVert \hat{g}_d + \rho_{\mathrm{eff}}\, \hat{g}_a \rVert < 10^{-3}$.
Given the gate, this cannot occur for any $\theta_{\mathrm{conf}}$
(the norm is bounded below by $0.5$ on
$\theta_{\mathrm{conf}} \le 150^\circ$), and it never triggered in
our runs.}
\begin{algorithm}[t]
\caption{GradBlend update}
\label{alg:gradblend}
\begin{algorithmic}[1]
\Require denoising gradient $g_d$, auxiliary gradient $g_a$
\Ensure blended update $g_{\mathrm{blend}}$
\State $\hat{g}_d \gets g_d / \lVert g_d \rVert$;\quad
       $\hat{g}_a \gets g_a / \lVert g_a \rVert$
\State $\theta_{\mathrm{conf}} \gets
       \arccos(\hat{g}_d \cdot \hat{g}_a)$
\State $m_{\mathrm{ref}} \gets \lVert g_d \rVert \cdot
       \max\!\left(\cos(\theta_{\mathrm{conf}}/2),\, 0.05\right)$
       \Comment{Eqs.~\eqref{eq:closed-form}--\eqref{eq:magnitude}}
\State $\rho_{\mathrm{eff}} \gets
  \begin{cases}
    1 & \theta_{\mathrm{conf}} \le 120^\circ \\
    1 - \frac{\theta_{\mathrm{conf}} - 120^\circ}{30^\circ}
        & 120^\circ < \theta_{\mathrm{conf}} \le 150^\circ \\
    0 & \theta_{\mathrm{conf}} > 150^\circ
  \end{cases}$
\State $\hat{g}_{\mathrm{dir}} \gets
       (\hat{g}_d + \rho_{\mathrm{eff}}\, \hat{g}_a) \,/\,
       \lVert \hat{g}_d + \rho_{\mathrm{eff}}\, \hat{g}_a \rVert$
\State $g_{\mathrm{blend}} \gets m_{\mathrm{ref}} \cdot
       \hat{g}_{\mathrm{dir}}$
\State \Return $g_{\mathrm{blend}}$
\end{algorithmic}
\end{algorithm}
Relative to a pure denoising step, the squared step-size ratio is
$R^2 = \max(\cos(\theta_{\mathrm{conf}}/2), 0.05)^2$, reaching one only
at perfect alignment (Appendix~\ref{app:diag-definitions}). The $150^\circ$
fallback and the denoising anchor together keep denoising primary.

\subsection{Auxiliary Weight Schedule}
\label{sec:schedule}
 
The schedule $w_{\mathrm{aux}}(s)$ decides when the auxiliary loss
is active. Because GradBlend normalizes each gradient to unit
length and anchors the step size to the denoising-gradient norm
(Section~\ref{sec:gradblend}), any positive value of $w_{\mathrm{aux}}(s)$ cancels from the update. The schedule is therefore a gate: the update is
pure denoising when $w_{\mathrm{aux}}(s) = 0$, and the auxiliary
enters at full unit weight otherwise. \textsc{Lantern} consequently
exposes no loss-weight hyperparameter; the only design choice is
when the gate opens and when it closes, which we study through
three gating windows in Section~\ref{sec:ablation}. We implement the gate with cosine ramps rather than a hard switch: the ramps set the scalar
auxiliary loss we report, and they are what the multi-task
baselines consume, since their updates scale with
$w_{\mathrm{aux}}(s)$ directly. Every method that uses the
auxiliary loss receives the same schedule, so differences between
methods reflect the combination rule and not the schedule.

\section{Experiments}
\label{sec:experiments}

\paragraph{\textbf{Dataset}}
All experiments use Dataset 2 of the Fast Calorimeter Simulation Challenge
(CaloChallenge)~\cite{faucci_giannelli_2022_6366271}, which contains electron
showers simulated with \textsc{Geant4} and incident energies drawn
log-uniformly between 1~GeV and 1~TeV. Each shower has $V = 6{,}480$ voxels over
45 layers, with 16 azimuthal ($\varphi$) and 9 radial bins per layer. The first
file of 100{,}000 showers is used for training and the second as the held-out
evaluation reference. Dataset 2 is a standard benchmark whose granularity is
representative of modern calorimeters, and our method carries over to the finer Dataset 3~\cite{faucci_giannelli_2022_6366324} from CaloChallenge 2022 geometry without modification.

\paragraph{\textbf{Experimental Setup}}
\label{sec:setup}
The surrogate is the two-stage model of Section~\ref{sec:architecture}: an energy network that generates the per-layer energy ratios $u$ and a shape network that generates the voxelized shower. The two auxiliary losses and GradBlend attach to the shape network as described there. This subsection fixes the training configuration.\footnote{Code is available at \url{https://github.com/Aaheer17/lantern-kdd27}.}

Both networks are denoising diffusion probabilistic models \cite{ho2020denoising} with 1{,}000 diffusion steps. Preprocessing follows the transform defined in Appendix~\ref{app:preproc}.

Training uses a single NVIDIA V100 GPU with 32~GB of memory for both networks. The energy network is trained once for 500 epochs in approximately one hour and is fixed across all configurations. The shape network trains for 800 epochs and dominates the overall training cost, taking approximately 38 hours for most gradient-based methods (ConFIG, GradBlend, IMTL-G, and PCGrad); GradNorm is somewhat more expensive at roughly 50 hours, owing to its additional per-task gradient-norm computation. All remaining hyperparameters are listed in Table~\ref{tab:hyperparams} of Appendix~\ref{app:arch}.

Three shape networks are trained independently with different random seeds. All results are reported as mean and standard deviation over the three seeds.


\subsection{Evaluation Metrics}
\label{sec:eval-metrics}

We report CFD (Section~\ref{sec:intro}) at two scales: layer-wise, over per-layer energies (Appendix~\ref{app:cfd-layer}), and voxel-wise, over the longitudinal structure between adjacent layers, defined next.

\paragraph{Voxel-wise CFD}
Voxel-wise CFD measures how well a model reproduces the longitudinal correlation
between a voxel and the voxel at the same transverse position in the following
layer. For each consecutive layer pair $(i, i{+}1)$ with $i = 1, \dots, 44$, and
each transverse position $(a, r)$ with angular bin $a = 1, \dots, 16$ and radial
bin $r = 1, \dots, 9$, we compute the Pearson correlation across showers between
the energy at $(i, a, r)$ and at $(i{+}1, a, r)$:
\begin{equation}
\rho^{(i)}_{a,r}
= \frac{\sum_k \big(E_{i,a,r,k} - \bar{E}_{i,a,r}\big)\big(E_{i+1,a,r,k} - \bar{E}_{i+1,a,r}\big)}
       {\sqrt{\sum_k \big(E_{i,a,r,k} - \bar{E}_{i,a,r}\big)^2}\;
        \sqrt{\sum_k \big(E_{i+1,a,r,k} - \bar{E}_{i+1,a,r}\big)^2}} ,
\label{eq:pearson}
\end{equation}
where $E_{i,a,r,k}$ is the energy at position $(a,r)$ in layer $i$ for shower
$k$, the index $k$ runs over the $K$ showers in the set, and
$\bar{E}_{i,a,r} = \tfrac{1}{K}\sum_k E_{i,a,r,k}$ is the mean energy at that
position across showers. The superscript $(i)$ identifies the layer pair
$(i, i{+}1)$, and the subscript $(a,r)$ the transverse position. Collecting
$\rho^{(i)}_{a,r}$ over all 44 layer pairs and all $16 \times 9$ transverse
positions gives a correlation tensor $\rho \in \mathbb{R}^{44 \times 16 \times 9}$
for \textsc{Geant4} ($\rho^\text{ref}$) and for the generated samples
($\rho^\text{gen}$). We evaluate only the active positions, defined as those
where both distributions carry nonzero variance, since the correlation is
undefined elsewhere. Voxel-wise CFD is the normalized Frobenius distance between
the two correlation tensors,
\begin{equation}
\mathrm{CFD}_\text{vox}
= \frac{\big\| \rho^\text{gen} - \rho^\text{ref} \big\|_F}
       {\big\| \rho^\text{ref} \big\|_F} ,
\qquad
\big\| \rho \big\|_F = \sqrt{\sum_{i,a,r} \big(\rho^{(i)}_{a,r}\big)^2}\, ,
\label{eq:cfd-vox}
\end{equation}
with both sums restricted to the active positions. Lower values indicate closer agreement with \textsc{Geant4}'s longitudinal correlation structure.

Alongside CFD and FPD/KPD, we report precision, recall, density,
and coverage (PRDC)~\cite{naeem2020reliable}, which separate sample quality from diversity, and the binary classifier test~\cite{krause2021caloflow_2}, which trains a network to distinguish generated from Geant4 showers on low-level (voxel) and high-level (physics) features; an AUC near 0.5 indicates
the two are hard to separate. Protocol details for all metrics in Appendix ~\ref{app:eval}.

\subsection{Baselines}
\label{sec:baselines}
We consider two groups of baselines. Prior surrogates compare the absolute quality of the generated showers; combination baselines isolate the gradient combination rule as the varying component.

The prior surrogates are CaloDiffusion \cite{amram2023calodiffusion}, CaloScore~v2 \cite{mikuni2024caloscore2}, CaloDREAM \cite{favaro2025calodream}, CaloDiT \cite{raikwar2024calodit} and CaloDiT-2~\cite{raikwar2025generalisable}. CaloDREAM uses conditional flow matching. Standard metrics come from the CaloChallenge report \cite{krause2025calochallenge}, which evaluated all submissions with the official pipeline against the same reference file used here. We compute CFD for all models with a single procedure: from the publicly released submission samples for the prior surrogates~\cite{faucci_giannelli_2025_15962050}, and from our own trained models for the combination baselines. The exception is CaloDiT-2, which postdates the CaloChallenge evaluation: its standard metrics come from its publication~\cite{raikwar2025generalisable} and its CFD is omitted (Appendix~\ref{app:reference}). 

The combination baselines hold the architecture, auxiliary losses, weight schedule, and preprocessing fixed, and vary only the gradient-combination rule. This group contains the denoising-only model, trained without auxiliary losses, and four combination methods: PCGrad \cite{yu2020pcgrad}, GradNorm \cite{chen2018gradnorm}, IMTL-G \cite{liu2021towards}, and ConFIG \cite{liu2025config}. Beyond these gradient-level rules, we also consider two basic loss-weighting schemes, weighted sum and uncertainty weighting~\cite{kendall2018multi}, which merge both objectives into one scalar loss and backpropagate once. Acting on the summed loss rather than individual gradients, they cannot address the gradient conflict this work targets, so we defer them to Appendix~\ref{app:wsum_uw_results}: plain weighted sum and unscheduled uncertainty weighting inflate FPD by orders of magnitude, and only a scheduled uncertainty variant, which reuses our weight schedule (Section~\ref{sec:schedule}), is competitive (Tables~\ref{tab:model_metrics_1} and~\ref{tab:model_metrics_prdc}).

\textsc{Base} denotes the denoising-only model. Models that add the auxiliary losses under a combination rule are written \textsc{Base+X}, for example \textsc{Base+PCGrad}. The proposed model, \textsc{Base+GradBlend}, is written \textsc{Lantern}. The prior surrogates keep their original names.

\subsection{Results and Discussion}
\label{sec:results-discussion}
\begin{table*}[t]
\centering
\caption{Shower fidelity on CaloChallenge Dataset~2. FPD and KPD are
scaled by $10^{3}$; lower is better for FPD, KPD, and
$\mathrm{CFD}_{\mathrm{vox}}$, and for AUC values closer to $0.5$ are
better. Classifier AUC is reported for high-level (HL) and low-level
(LL) features. $\mathrm{CFD}_{\mathrm{vox}}$ is the voxel-wise
Correlation Frobenius Distance of Eq.~\eqref{eq:cfd-vox}, computed in
this work for all models including the prior surrogates; the layer-wise
variant is defined in Appendix~\ref{app:cfd-layer}. Remaining
standard metrics for the prior surrogates are taken from the
CaloChallenge report~\cite{krause2025calochallenge}. All auxiliary-loss
models are trained under schedule~S3 (Section~\ref{sec:ablation});
differences reflect the combination rule alone. Trained models report
mean~$\pm$~std over three seeds, and the best value among the trained
models per column is in bold.}
\label{tab:main}
\resizebox{0.85\textwidth}{!}{%
\begin{tabular}{llccccc}
\toprule
Model & Loss & FPD $\downarrow$ & KPD $\downarrow$ & $\mathrm{CFD}_{\mathrm{vox}}$ $\downarrow$ & AUC (HL) & AUC (LL) \\
\midrule
CaloDiffusion \cite{amram2023calodiffusion} & N/A & $146.9 \pm 0.9$ & $0.174 \pm 0.042$ & $0.155$ & $0.591 \pm 0.009$ & $0.577 \pm 0.004$ \\
CaloScore~v2 \cite{mikuni2024caloscore2}    & N/A & $112.5 \pm 0.9$ & $0.149 \pm 0.057$ & $0.176$ & $0.666 \pm 0.002$ & $0.595 \pm 0.003$ \\
CaloDiT \cite{raikwar2024calodit}           & N/A & $1691 \pm 7$ & $11.030 \pm 0.427$ & $0.979$ & $0.912 \pm 0.002$ & $0.984 \pm 0.001$ \\
CaloDiT-2 (EDM) \cite{raikwar2025generalisable} & N/A & $20.06 \pm 0.74$ & $0.089 \pm 0.090$ & -- & $0.560 \pm 0.005$ & $0.594 \pm 0.002$ \\
CaloDiT-2 (CD) \cite{raikwar2025generalisable}  & N/A & $68.83 \pm 3.37$ & $0.390 \pm 0.150$ & -- & $0.695 \pm 0.004$ & $0.598 \pm 0.002$ \\
CaloDREAM \cite{favaro2025calodream}        & N/A & $24.65 \pm 1.04$ & $0.023 \pm 0.036$ & $0.113$ & $0.521 \pm 0.002$ & $0.531 \pm 0.003$ \\
\midrule
\base                & None     & $30.87 \pm 1.26$ & $0.002 \pm 0.001$  & $0.122 \pm 0.004$ & $\mathbf{0.531 \pm 0.009}$ & $\mathbf{0.545 \pm 0.013}$ \\
\midrule
\basex{PCGrad}       & Voxel     & $68.50 \pm 44.58$ & $0.137 \pm 0.191$ & $0.142 \pm 0.008$ & $0.566 \pm 0.027$ & $0.560 \pm 0.003$ \\
\basex{GradNorm}     & Voxel     & $60.27 \pm 11.15$ & $0.102 \pm 0.042$ & $0.142 \pm 0.014$ & $0.549 \pm 0.021$ & $0.563 \pm 0.001$ \\
\basex{IMTL-G}       & Voxel     & $406.9 \pm 80.9$ & $2.216 \pm 0.538$ & $0.192 \pm 0.012$ & $0.807 \pm 0.019$ & $0.763 \pm 0.011$ \\
\basex{ConFIG}       & Voxel     & $204.7 \pm 31.2$ & $0.882 \pm 0.110$ & $0.156 \pm 0.009$ & $0.636 \pm 0.003$ & $0.592 \pm 0.002$ \\
\lantern             & Voxel    & $32.84 \pm 3.27$ & $-0.0007 \pm 0.0046$  & $0.125 \pm 0.005$ & $0.538 \pm 0.002$ & $0.558 \pm 0.002$ \\
\midrule
\basex{PCGrad}       & Laplacian & $169.0 \pm 72.8$ & $0.640 \pm 0.392$ & $0.155 \pm 0.003$ & $0.621 \pm 0.033$ & $0.580 \pm 0.021$ \\
\basex{GradNorm}     & Laplacian & $133.5 \pm 51.5$ & $0.472 \pm 0.298$ & $0.147 \pm 0.011$ & $0.620 \pm 0.036$ & $0.582 \pm 0.005$ \\
\basex{IMTL-G}       & Laplacian & $3454 \pm 4820$ & $23.520 \pm 34.500$ & $0.510 \pm 0.313$ & $0.874 \pm 0.106$ & $0.890 \pm 0.088$ \\
\basex{ConFIG}       & Laplacian & $262.8 \pm 73.9$ & $1.209 \pm 0.341$ & $0.172 \pm 0.007$ & $0.656 \pm 0.033$ & $0.623 \pm 0.015$ \\
\lantern             & Laplacian & $\mathbf{25.97 \pm 1.44}$ & $-0.0048 \pm 0.0009$ & $\mathbf{0.120 \pm 0.005}$ & $0.534 \pm 0.003$ & $0.554 \pm 0.002$ \\
\bottomrule
\end{tabular}
}
\end{table*}

Table~\ref{tab:main} reports shower fidelity on Dataset~2. Among the
prior surrogates, CaloDiffusion, CaloScore~v2, and the original CaloDiT
trail on FPD and CFD, CaloDiT severely so. CaloDiT-2 (EDM) reports the
lowest FPD in the table, its single-step CD variant trading fidelity for
sampling speed, but both show weaker classifier AUCs than CaloDREAM and
their CFD cannot be assessed (Section~\ref{sec:baselines}). CaloDREAM,
the flow-matching model our two-stage architecture adapts, attains the
best CFD and classifier AUCs and serves as our reference; \lantern
reaches comparable FPD and CFD with a DDPM backbone.

\base is a strong baseline in its own right, reaching an FPD of $30.9$
and a CFD of $0.1215$ without any auxiliary loss. This sets a high bar:
an auxiliary physics loss is useful only if it can be added without
disturbing this baseline.

The general-purpose multi-task gradient methods do not meet that bar.
Every \basex{PCGrad}, \basex{GradNorm}, \basex{IMTL-G}, and
\basex{ConFIG} row has a higher FPD than \base, in several cases by more
than an order of magnitude, with standard deviations large relative to
the means. \basex{IMTL-G} is the worst, diverging on at least one seed
under the Laplacian loss. These are unstable outcomes, not small
regressions.

\begin{figure}[t]
  \centering
  \begin{subfigure}{\columnwidth}
    \includegraphics[width=\linewidth]{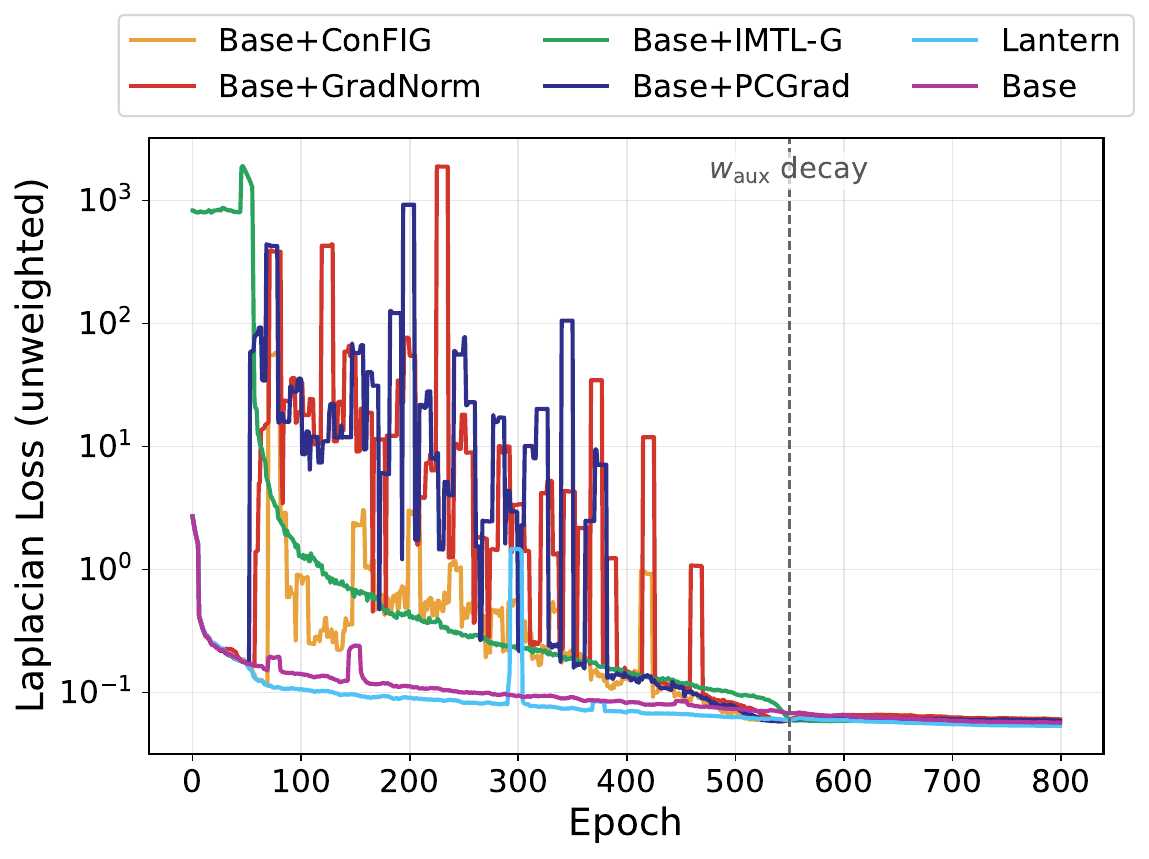}
    \caption{Laplacian loss.}
    \label{fig:laplacian_loss}
  \end{subfigure}\\[2pt]
  \begin{subfigure}{\columnwidth}
    \includegraphics[width=\linewidth]{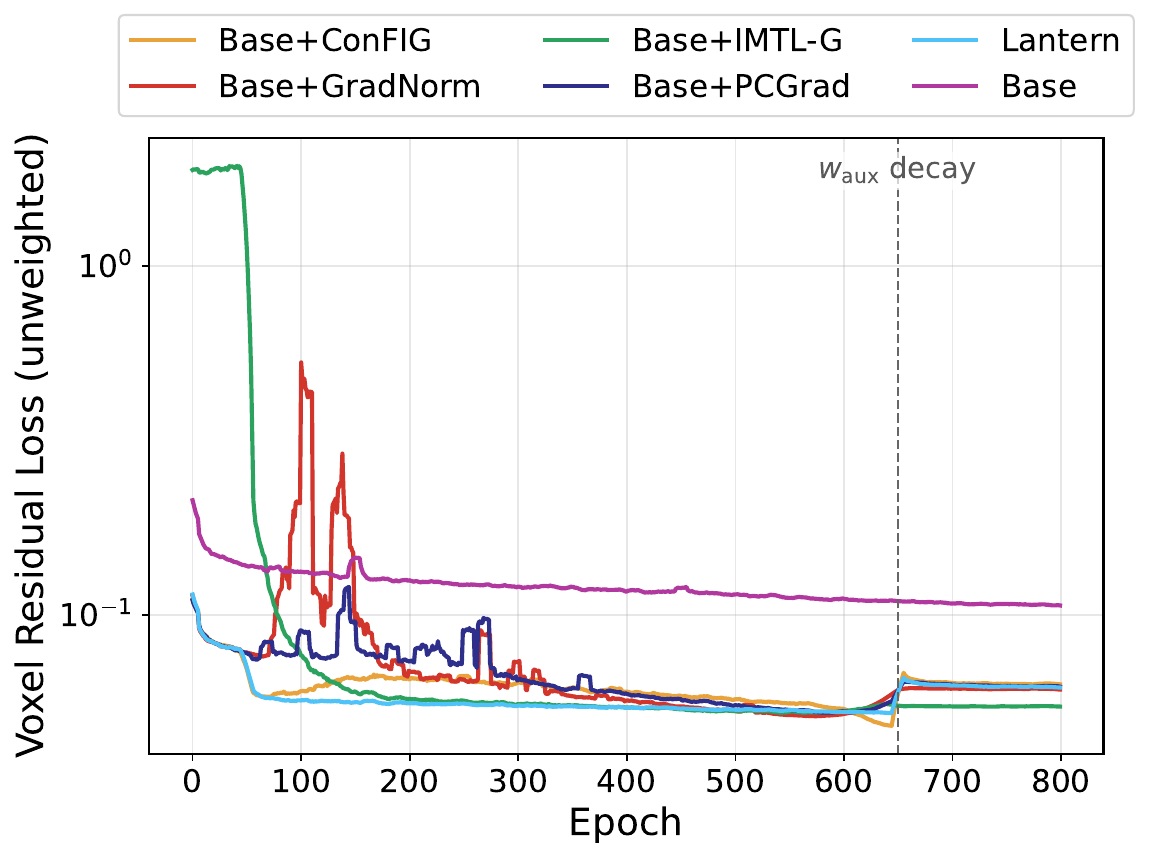}
    \caption{Voxel residual loss.}
    \label{fig:voxel_loss}
  \end{subfigure}
  \caption{Unweighted auxiliary loss over training on Dataset~2. The
  general-purpose multi-task gradient methods produce large fluctuations, rising
  by several orders of magnitude at points during training. GradBlend
  (\lantern) keeps the auxiliary loss stable, close to the \base curve that
  never optimizes it. The dashed line marks the epoch at which the auxiliary
  weight begins to decay.}
  \label{fig:aux_loss_curves}
 \vspace{-8pt} 
\end{figure}

The gradient diagnostics explain the pattern. Because these methods treat
the denoising and auxiliary gradients as co-equal, the combined update
drifts from the denoising direction that produces valid samples, degrading
fidelity; Section~\ref{sec:diagnostics} reads the per-method allocations
from Table~\ref{tab:rq4:ratios}. The training curves in
Figure~\ref{fig:aux_loss_curves} show the resulting instability, the
auxiliary loss rising by orders of magnitude and strongest under the
Laplacian loss, whose high-pass gradient is noisier.
\lantern is the only combination method that adds the auxiliary loss
without this damage. Under the Laplacian loss it reaches the best FPD and
CFD among the trained models, slightly ahead of \base; under the voxel
loss it stays close to \base, with small standard deviations. \lantern
integrates the auxiliary objective and preserves the fidelity of the
denoising-only model.
KPD does not separate the strongest models: \lantern and \base are both
near zero, \lantern slightly negative. KPD is an unbiased MMD estimate
that can go slightly negative when samples closely match the reference,
so this is a close match, not an error, while the failing methods have
large positive KPD consistent with their FPD. Classifier AUCs in the
\base and \lantern rows are all near $0.5$. Precision, recall, density,
and coverage confirm the failure mode: the multi-task baselines gain
precision and density but collapse in recall, the signature of mode
collapse, whereas \base and \lantern stay balanced and close to the
reference-quality prior surrogates (full breakdown in
Appendix~\ref{app:prdc-schedule}, Table~\ref{tab:prdc}).


%

\subsection{Ablation Studies}
\label{sec:ablation}
We compare three gating windows, holding every other setting fixed.
Schedule~S1 opens the gate at the first optimization step. S2 opens it
after a 50-epoch warmup, reaching full weight by epoch~250 and holding
to the end. S3 matches S2 but closes the gate before the end, decaying
to zero by epoch~650 for the voxel loss and~550 for the Laplacian loss;
the closing epochs were fixed a priori, not tuned. The comparison is
factored: S1 versus S2 tests whether denoising must settle before the
auxiliary enters, S2 versus S3 whether it must finish alone.

Timing decides voxel sample quality. Held to the last epoch (S1, S2),
the voxel loss inflates FPD and density far above one while recall
collapses (Table~\ref{tab:schedule-ablation}; full PRDC in
Appendix~\ref{app:prdc-schedule}): the generator concentrates mass on
part of the real distribution and fails to cover it. Coverage stays
near~$0.99$ throughout, so this is a recall collapse, not a coverage
failure. S3 reverses it, returning FPD, density, and the AUCs to near
their calibrated values; precision drops correspondingly, so the
terminal gate trades a little per-sample fidelity for a large gain in
coverage and calibration.

The Laplacian loss, by contrast, is insensitive to the schedule: FPD
spans only $23.5$ to $26.0$ and $\mathrm{CFD}_{\mathrm{vox}}$ $0.118$ to
$0.129$ across S1 through S3, and its precision, density, recall, and
coverage agree across schedules to within run-to-run variation
(Table~\ref{tab:schedule-ablation-prdc}). The voxel loss competes with
denoising late in training while the Laplacian loss does not, so gating
it leaves the optimization trajectory and the samples unchanged. The
gate is therefore necessary where the auxiliary conflicts with denoising
and inert where it does not. \lantern{} adopts S3 for both losses,
removing the voxel conflict in the final epochs at no cost to the
Laplacian loss; this keeps a single schedule across auxiliaries and
avoids per-loss tuning.

\begin{table}[t]
\centering
\caption{Effect of the auxiliary loss schedule on \lantern.}
\label{tab:schedule-ablation}
\small
\setlength{\tabcolsep}{2.5pt}
\begin{tabular}{llccccc}
\toprule
Auxiliary & Sched. & FPD $\downarrow$ & KPD $\downarrow$ & $\mathrm{CFD}_{\mathrm{vox}}$ $\downarrow$ & AUC$_{\mathrm{HL}}$ & AUC$_{\mathrm{LL}}$ \\
\midrule
\multirow{3}{*}{Voxel}
  & S1 & $173.4$ & $0.997$ & $0.123$ & $0.777$ & $0.783$ \\
  & S2 & $236.3$ & $1.528$ & $0.138$ & $0.833$ & $0.838$ \\
  & S3 & $32.84$ & $-0.0007$ & $0.125$ & $0.538$ & $0.558$ \\
\addlinespace[4pt]
\multirow{3}{*}{Laplacian}
  & S1 & $25.34$ & $0.001$ & $\mathbf{0.118}$ & $0.534$ & $0.561$ \\
  & S2 & $\mathbf{23.53}$ & $0.001$ & $0.129$ & $0.536$ & $0.560$ \\
  & S3 & $25.97$ & $-0.0048$ & $0.120$ & $0.534$ & $0.554$ \\
\bottomrule
\end{tabular}
\vspace{-8pt} 
\end{table}


\begin{table*}[!t]
\centering
\caption{Gradient diagnostics over the active window, voxel energy loss on
Dataset~2, reported as mean $\pm$ std over three seeds. $\theta_{\mathrm{conf}}$
is the conflict angle between the denoising gradient $g_d$ and the auxiliary
gradient $g_a$; $R_1^{d}$ and $R_1^{a}$ are first-order progress ratios of the
blended step $g_{\mathrm{blend}}$ on the two objectives; Balance is
$R_1^{d}/R_1^{a}$, where a value above one favors the denoising objective;
$\cos_{d}$ and $\cos_{a}$ are the cosines of $g_{\mathrm{blend}}$ with $g_d$ and
$g_a$; \%spike is the fraction of active epochs removed by the auxiliary-norm
spike filter. Bold marks the best value among the combination methods per
column.}
\label{tab:rq4:ratios}
  \resizebox{0.8\textwidth}{!}{%
\begin{tabular}{lrrrrrrr}
\toprule
Method & $\theta_{\mathrm{conf}}$ & $R_1^{d}$ & $R_1^{a}$ & Balance & $\cos_{d}$ & $\cos_{a}$ & \%spike \\
\midrule
\basex{ConFIG}   & $92.0 \pm 1.2$  & $1.42 \pm 0.05$ & $0.78 \pm 0.01$ & $1.81 \pm 0.05$ & $0.69 \pm 0.01$ & $0.69 \pm 0.01$ & $12 \pm 0$ \\
\basex{GradNorm} & $127.5 \pm 3.6$ & $0.55 \pm 0.09$ & $0.66 \pm 0.07$ & $0.83 \pm 0.05$ & $0.36 \pm 0.02$ & $0.41 \pm 0.03$ & $9 \pm 0$ \\
\basex{IMTL-G}   & $116.7 \pm 3.3$ & $0.14 \pm 0.01$ & $0.10 \pm 0.01$ & $1.33 \pm 0.11$ & $0.52 \pm 0.02$ & $0.52 \pm 0.02$ & $13 \pm 1$ \\
\basex{PCGrad}   & $123.9 \pm 7.1$ & $0.95 \pm 0.17$ & $0.44 \pm 0.05$ & $2.13 \pm 0.20$ & $0.55 \pm 0.08$ & $0.31 \pm 0.02$ & $11 \pm 0$ \\
\lantern         & $119.7 \pm 0.5$ & $0.71 \pm 0.02$ & $0.24 \pm 0.07$ & $\mathbf{3.26 \pm 1.16}$ & $\mathbf{0.71 \pm 0.02}$ & $\mathbf{0.21 \pm 0.04}$ & $\mathbf{8 \pm 0}$ \\
\bottomrule
\end{tabular}
}
\vspace{-8pt} 
\end{table*}

\subsection{Training Diagnostics}
\label{sec:diagnostics}

Table~\ref{tab:rq4:ratios} reports gradient diagnostics for the voxel energy
loss, measured over the active window and averaged over three seeds. Each row
summarizes how a combination method allocates its update between the denoising
and auxiliary objectives. Definitions of the ratios, the active window, and the
spike filter are given in Appendix~\ref{app:diag-definitions}.



The two cosine columns separate the methods. For \basex{ConFIG} and
\basex{IMTL-G}, $\cos_{d}$ equals $\cos_{a}$ in every seed. These methods weight the two gradients equally, so the resulting step makes the same angle with each. \lantern shows the widest gap between the two cosines, $0.71$ against $0.21$. Its step stays aligned with the denoising gradient and is held away from the auxiliary gradient. \basex{PCGrad} shows a partial gap, \basex{GradNorm} a small one.

The Balance column reads which objective each step favors. \lantern keeps the
denoising objective ahead in every seed, with Balance above two.
\basex{GradNorm} inverts this order and favors the auxiliary gradient over the
denoiser, with Balance below one.

A large $R_1^{d}$ is not a measure of sample quality. \basex{ConFIG} reaches
$R_1^{d} = 1.42$, the highest in the table, yet has poor FPD
(Table~\ref{tab:main}). The ratios describe how the step is allocated and should
be read together with the evaluation tables.

The spike column reports the fraction of active epochs removed by the
auxiliary-norm spike filter. \lantern has the lowest rate at eight percent, and
\basex{IMTL-G} the highest. This ordering matches the stability seen in the
training curves (Figure~\ref{fig:aux_loss_curves}).


Figure~\ref{fig:aux_loss_curves} shows the unweighted auxiliary loss over
training. The general-purpose multi-task methods fluctuate by orders of
magnitude as the auxiliary gradient periodically overtakes the update, whereas
\lantern keeps it smooth, matching the \base curve that never optimizes it while
still carrying the auxiliary signal.

\section{Conclusion}
\label{sec:conclusion}
Physics-informed diffusion typically relies on a closed-form physical law to supervise generation, which calorimeter shower simulation lacks. We presented \lantern{}, which guides a diffusion surrogate with two per-sample physics losses, a variance-stabilized voxel energy residual and a graph Laplacian smoothness term, and merges their gradients with the denoising gradient through GradBlend, a combination rule that keeps denoising as the trusted reference direction while admitting each physics gradient where it agrees. These losses and GradBlend apply to the second stage, which predicts voxel energies conditioned on the first-stage incident energy and layer ratios. To measure fidelity FPD and KPD miss, we introduced CFD, the Correlation Frobenius Distance, which scores the correlation structure of calorimeter showers at two scales. On CaloChallenge Dataset~2, \lantern{} integrates both physics losses without the degradation symmetric multi-task rules incur: PCGrad, GradNorm, IMTL-G, and ConFIG raise FPD by factors of two to over one hundred, whereas \lantern{} stays close to the denoising-only baseline on FPD and KPD and remains competitive with the strongest prior surrogates on CFD and classifier AUC.
We treat denoising as the primary objective; the ablations show that an auxiliary loss strengthens the model as long as its gradient stays aligned with it. The voxel gradient turns from cooperative to opposing over training, so holding it at full weight to the final epoch inflates FPD and collapses recall; a terminal phase that trains on the primary loss alone removes this and recovers both. The Laplacian gradient stays aligned throughout training and is insensitive to the schedule, so a single schedule serves both losses without per-loss tuning. We discuss limitations and next steps in Section~\ref{sec:limitations}.

\section{Limitations and Ethical Considerations}
\label{sec:limitations}


\paragraph{Limitations.}
GradBlend and the two physics losses are demonstrated on a single controlled setting; transfer to other configurations remains future work. The geometries released with LEMURS~\cite{raikwar2025generalisable}, the higher-granularity Dataset~3, and hadronic showers are direct targets: adapting the method to them is an application, not a reformulation. The approach carries two sets of empirically set thresholds. The weight schedule's closing epochs, and GradBlend's angular thresholds ($120^\circ$ for decay and $150^\circ$ for the fallback) together with the magnitude floor ($0.05$), were stable across the voxel and Laplacian losses here, but neither is derived, and both the schedule and the direction rule warrant further analysis, ideally tied to the measured conflict angle. We study each auxiliary loss on its own and do not characterize how the voxel-residual and Laplacian objectives interact when trained together. Results are reported over three random seeds of the shape network with the energy network fixed, so the reported variance reflects shape-network initialization alone; three seeds is common but limits power to resolve small metric differences. Physics guidance acts entirely during training and leaves the sampler unchanged, so fidelity gains come at no added generation cost. Reducing the base surrogate's sampling cost is orthogonal to our contributions and can be combined with them.
\paragraph{Ethical considerations.}
This work develops a surrogate model for calorimeter shower simulation in high-energy physics, a fundamental-physics application trained and evaluated entirely on simulated detector data from Geant4, with no human subjects, personal or sensitive data, or application to individuals. The intended benefit is reduced computational cost for physics simulation, lowering the energy and hardware demands of large-scale Monte Carlo production. Generated samples approximate rather than replace full simulation, as with any surrogate, so downstream physics analyses using them should account for the residual modeling error characterized here, particularly the correlation-structure error CFD is designed to expose. Calorimeter shower generation has no foreseeable harmful dual-use application beyond its intended scientific scope.

\section{Generative AI Usage}
\label{sec:genai}

A large language model assisted with manuscript editing and LaTeX formatting, portions of the experiment code, and cross-checking equations and related work. All contributions, including the method, losses, metric, and experimental design, are the authors' own; the authors have reviewed all AI-assisted material and take full responsibility for the paper's content, correctness, and originality.
\clearpage
\bibliographystyle{ACM-Reference-Format}
\bibliography{sample-base}


\begin{thebibliography}{52}


\ifx \showCODEN    \undefined \def \showCODEN     #1{\unskip}     \fi
\ifx \showISBNx    \undefined \def \showISBNx     #1{\unskip}     \fi
\ifx \showISBNxiii \undefined \def \showISBNxiii  #1{\unskip}     \fi
\ifx \showISSN     \undefined \def \showISSN      #1{\unskip}     \fi
\ifx \showLCCN     \undefined \def \showLCCN      #1{\unskip}     \fi
\ifx \shownote     \undefined \def \shownote      #1{#1}          \fi
\ifx \showarticletitle \undefined \def \showarticletitle #1{#1}   \fi
\ifx \showURL      \undefined \def \showURL       {\relax}        \fi
\providecommand\bibfield[2]{#2}
\providecommand\bibinfo[2]{#2}
\providecommand\natexlab[1]{#1}
\providecommand\showeprint[2][]{arXiv:#2}

\bibitem[Agostinelli et~al\mbox{.}(2003)]%
        {agostinelli2003geant4}
\bibfield{author}{\bibinfo{person}{S. Agostinelli} {et~al\mbox{.}}}
  \bibinfo{year}{2003}\natexlab{}.
\newblock \showarticletitle{{GEANT4}---A Simulation Toolkit}.
\newblock \bibinfo{journal}{\emph{Nuclear Instruments and Methods in Physics
  Research A}} \bibinfo{volume}{506}, \bibinfo{number}{3}
  (\bibinfo{year}{2003}), \bibinfo{pages}{250--303}.
\newblock


\bibitem[Ahmad et~al\mbox{.}(2024)]%
        {ahmad2024comprehensive}
\bibfield{author}{\bibinfo{person}{Farzana~Yasmin Ahmad},
  \bibinfo{person}{Vanamala Venkataswamy}, {and} \bibinfo{person}{Geoffrey
  Fox}.} \bibinfo{year}{2024}\natexlab{}.
\newblock \showarticletitle{A comprehensive evaluation of generative models in
  calorimeter shower simulation}.
\newblock \bibinfo{journal}{\emph{arXiv preprint arXiv:2406.12898}}
  (\bibinfo{year}{2024}).
\newblock


\bibitem[Amram and Pedro(2023)]%
        {amram2023calodiffusion}
\bibfield{author}{\bibinfo{person}{Oz Amram} {and} \bibinfo{person}{Kevin
  Pedro}.} \bibinfo{year}{2023}\natexlab{}.
\newblock \showarticletitle{Denoising Diffusion Models with Geometry Adaptation
  for High Fidelity Calorimeter Simulation}.
\newblock \bibinfo{journal}{\emph{Physical Review D}}  \bibinfo{volume}{108}
  (\bibinfo{year}{2023}), \bibinfo{pages}{072014}.
\newblock


\bibitem[Apollinari et~al\mbox{.}(2017)]%
        {apollinari2017high}
\bibfield{author}{\bibinfo{person}{Giorgio Apollinari}, \bibinfo{person}{O
  Br{\"u}ning}, \bibinfo{person}{Tatsushi Nakamoto}, {and}
  \bibinfo{person}{Lucio Rossi}.} \bibinfo{year}{2017}\natexlab{}.
\newblock \showarticletitle{High luminosity large hadron collider HL-LHC}.
\newblock \bibinfo{journal}{\emph{arXiv preprint arXiv:1705.08830}}
  (\bibinfo{year}{2017}).
\newblock


\bibitem[Baldan et~al\mbox{.}(2025)]%
        {baldan2025flow}
\bibfield{author}{\bibinfo{person}{Giacomo Baldan}, \bibinfo{person}{Qiang
  Liu}, \bibinfo{person}{Alberto Guardone}, {and} \bibinfo{person}{Nils
  Thuerey}.} \bibinfo{year}{2025}\natexlab{}.
\newblock \showarticletitle{Flow matching meets pdes: A unified framework for
  physics-constrained generation}.
\newblock \bibinfo{journal}{\emph{arXiv preprint arXiv:2506.08604}}
  (\bibinfo{year}{2025}).
\newblock


\bibitem[Bastek et~al\mbox{.}(2025)]%
        {bastek2025pidm}
\bibfield{author}{\bibinfo{person}{Jan-Hendrik Bastek},
  \bibinfo{person}{WaiChing Sun}, {and} \bibinfo{person}{Dennis~M. Kochmann}.}
  \bibinfo{year}{2025}\natexlab{}.
\newblock \showarticletitle{Physics-Informed Diffusion Models}. In
  \bibinfo{booktitle}{\emph{International Conference on Learning
  Representations (ICLR)}}.
\newblock
\showeprint[arxiv]{2403.14404}~[cs.LG]


\bibitem[Birk et~al\mbox{.}(2024)]%
        {birk2024omnijet}
\bibfield{author}{\bibinfo{person}{Joschka Birk}, \bibinfo{person}{Anna
  Hallin}, {and} \bibinfo{person}{Gregor Kasieczka}.}
  \bibinfo{year}{2024}\natexlab{}.
\newblock \showarticletitle{{OmniJet-$\alpha$}: The first cross-task foundation
  model for particle physics}.
\newblock \bibinfo{journal}{\emph{Machine Learning: Science and Technology}}
  \bibinfo{volume}{5}, \bibinfo{number}{3} (\bibinfo{year}{2024}),
  \bibinfo{pages}{035031}.
\newblock
\href{https://doi.org/10.1088/2632-2153/ad66ad}{doi:\nolinkurl{10.1088/2632-2153/ad66ad}}


\bibitem[Buhmann et~al\mbox{.}(2024)]%
        {buhmann2024caloclouds2}
\bibfield{author}{\bibinfo{person}{Erik Buhmann}, \bibinfo{person}{Frank
  Gaede}, \bibinfo{person}{Gregor Kasieczka}, \bibinfo{person}{Anatolii Korol},
  \bibinfo{person}{William Korcari}, \bibinfo{person}{Katja Kr{\"u}ger}, {and}
  \bibinfo{person}{Peter McKeown}.} \bibinfo{year}{2024}\natexlab{}.
\newblock \showarticletitle{{CaloClouds II}: Ultra-Fast Geometry-Independent
  Highly-Granular Calorimeter Simulation}.
\newblock \bibinfo{journal}{\emph{Journal of Instrumentation}}
  \bibinfo{volume}{19} (\bibinfo{year}{2024}), \bibinfo{pages}{P04020}.
\newblock


\bibitem[{CaloChallenge Collaboration}(2022)]%
        {calochallenge_homepage}
\bibfield{author}{\bibinfo{person}{{CaloChallenge Collaboration}}.}
  \bibinfo{year}{2022}\natexlab{}.
\newblock \bibinfo{title}{{Fast Calorimeter Simulation Challenge 2022 --
  Homepage and Evaluation Code}}.
\newblock
  \bibinfo{howpublished}{\url{https://github.com/CaloChallenge/homepage}}.
\newblock
\shownote{Accessed: 2026-07-13}.
\newblock


\bibitem[Chen et~al\mbox{.}(2018)]%
        {chen2018gradnorm}
\bibfield{author}{\bibinfo{person}{Zhao Chen}, \bibinfo{person}{Vijay
  Badrinarayanan}, \bibinfo{person}{Chen-Yu Lee}, {and} \bibinfo{person}{Andrew
  Rabinovich}.} \bibinfo{year}{2018}\natexlab{}.
\newblock \showarticletitle{{GradNorm}: Gradient Normalization for Adaptive
  Loss Balancing in Deep Multitask Networks}. In
  \bibinfo{booktitle}{\emph{Proceedings of the 35th International Conference on
  Machine Learning (ICML)}} \emph{(\bibinfo{series}{Proceedings of Machine
  Learning Research}, Vol.~\bibinfo{volume}{80})}. \bibinfo{publisher}{PMLR},
  \bibinfo{pages}{794--803}.
\newblock


\bibitem[Chung et~al\mbox{.}(2023)]%
        {chung2023dps}
\bibfield{author}{\bibinfo{person}{Hyungjin Chung}, \bibinfo{person}{Jeongsol
  Kim}, \bibinfo{person}{Michael~T. McCann}, \bibinfo{person}{Marc~L. Klasky},
  {and} \bibinfo{person}{Jong~Chul Ye}.} \bibinfo{year}{2023}\natexlab{}.
\newblock \showarticletitle{Diffusion Posterior Sampling for General Noisy
  Inverse Problems}. In \bibinfo{booktitle}{\emph{International Conference on
  Learning Representations (ICLR)}}.
\newblock


\bibitem[Dosovitskiy et~al\mbox{.}(2020)]%
        {dosovitskiy2020image}
\bibfield{author}{\bibinfo{person}{Alexey Dosovitskiy}, \bibinfo{person}{Lucas
  Beyer}, \bibinfo{person}{Alexander Kolesnikov}, \bibinfo{person}{Dirk
  Weissenborn}, \bibinfo{person}{Xiaohua Zhai}, \bibinfo{person}{Thomas
  Unterthiner}, \bibinfo{person}{Mostafa Dehghani}, \bibinfo{person}{Matthias
  Minderer}, \bibinfo{person}{Georg Heigold}, \bibinfo{person}{Sylvain Gelly},
  {et~al\mbox{.}}} \bibinfo{year}{2020}\natexlab{}.
\newblock \showarticletitle{An image is worth 16x16 words: Transformers for
  image recognition at scale}.
\newblock \bibinfo{journal}{\emph{arXiv preprint arXiv:2010.11929}}
  (\bibinfo{year}{2020}).
\newblock


\bibitem[Du et~al\mbox{.}(2018)]%
        {du2018gcs}
\bibfield{author}{\bibinfo{person}{Yunshu Du}, \bibinfo{person}{Wojciech~M.
  Czarnecki}, \bibinfo{person}{Siddhant~M. Jayakumar}, \bibinfo{person}{Mehrdad
  Farajtabar}, \bibinfo{person}{Razvan Pascanu}, {and} \bibinfo{person}{Balaji
  Lakshminarayanan}.} \bibinfo{year}{2018}\natexlab{}.
\newblock \showarticletitle{Adapting Auxiliary Losses Using Gradient
  Similarity}.
\newblock
\shownote{arXiv:1812.02224}.
\newblock


\bibitem[Fabjan and Gianotti(2003)]%
        {fabjan2003calorimetry}
\bibfield{author}{\bibinfo{person}{Christian~W Fabjan} {and}
  \bibinfo{person}{Fabiola Gianotti}.} \bibinfo{year}{2003}\natexlab{}.
\newblock \showarticletitle{Calorimetry for particle physics}.
\newblock \bibinfo{journal}{\emph{Reviews of Modern Physics}}
  \bibinfo{volume}{75}, \bibinfo{number}{4} (\bibinfo{year}{2003}),
  \bibinfo{pages}{1243}.
\newblock


\bibitem[Faucci~Giannelli et~al\mbox{.}(2022a)]%
        {faucci_giannelli_2022_6366271}
\bibfield{author}{\bibinfo{person}{Michele Faucci~Giannelli},
  \bibinfo{person}{Gregor Kasieczka}, \bibinfo{person}{Claudius Krause},
  \bibinfo{person}{Ben Nachman}, \bibinfo{person}{Dalila Salamani},
  \bibinfo{person}{David Shih}, {and} \bibinfo{person}{Anna Zaborowska}.}
  \bibinfo{year}{2022}\natexlab{a}.
\newblock \bibinfo{booktitle}{\emph{Fast Calorimeter Simulation Challenge 2022
  - Dataset 2}}.
\newblock
\href{https://doi.org/10.5281/zenodo.6366271}{doi:\nolinkurl{10.5281/zenodo.6366271}}


\bibitem[Faucci~Giannelli et~al\mbox{.}(2022b)]%
        {faucci_giannelli_2022_6366324}
\bibfield{author}{\bibinfo{person}{Michele Faucci~Giannelli},
  \bibinfo{person}{Gregor Kasieczka}, \bibinfo{person}{Claudius Krause},
  \bibinfo{person}{Ben Nachman}, \bibinfo{person}{Dalila Salamani},
  \bibinfo{person}{David Shih}, {and} \bibinfo{person}{Anna Zaborowska}.}
  \bibinfo{year}{2022}\natexlab{b}.
\newblock \bibinfo{booktitle}{\emph{Fast Calorimeter Simulation Challenge 2022
  - Dataset 3}}.
\newblock
\href{https://doi.org/10.5281/zenodo.6366324}{doi:\nolinkurl{10.5281/zenodo.6366324}}


\bibitem[Faucci~Giannelli et~al\mbox{.}(2025)]%
        {faucci_giannelli_2025_15962050}
\bibfield{author}{\bibinfo{person}{Michele Faucci~Giannelli},
  \bibinfo{person}{Gregor Kasieczka}, \bibinfo{person}{Claudius Krause},
  \bibinfo{person}{Benjamin Nachman}, \bibinfo{person}{Dalila Salamani},
  \bibinfo{person}{David Shih}, \bibinfo{person}{Anna Zaborowska},
  \bibinfo{person}{Oz Amram}, \bibinfo{person}{Kerstin Borras},
  \bibinfo{person}{Matthew Buckley}, \bibinfo{person}{Thorsten Buss},
  \bibinfo{person}{Renato~Paulo Da~Costa~Cardoso}, \bibinfo{person}{Vijay
  Ekambaram}, \bibinfo{person}{Florian Ernst}, \bibinfo{person}{Luigi Favaro},
  \bibinfo{person}{Frank Gaede}, \bibinfo{person}{Shih-Chieh Hsu},
  \bibinfo{person}{Kristina Jaruskova}, \bibinfo{person}{Benno Käch},
  \bibinfo{person}{Jayant Kalagnanam}, \bibinfo{person}{Dirk Krücker},
  \bibinfo{person}{Qibin Liu}, \bibinfo{person}{Xiulong Liu},
  \bibinfo{person}{Thandikire Madula}, \bibinfo{person}{Isabell-Alissandra
  Melzer-Pellmann}, \bibinfo{person}{Vinicius Mikuni}, \bibinfo{person}{Nam
  Nguyen}, \bibinfo{person}{Ayodele Ore}, \bibinfo{person}{Sofia
  Palacios~Schweitzer}, \bibinfo{person}{Ian Pang}, \bibinfo{person}{Kevin
  Pedro}, \bibinfo{person}{Tilman Plehn}, \bibinfo{person}{Piyush Raikwar},
  \bibinfo{person}{John Raine}, \bibinfo{person}{Moritz Alfons~Wilhelm Scham},
  \bibinfo{person}{Simon Schnake}, \bibinfo{person}{Chase Shimmin},
  \bibinfo{person}{Eli Shlizerman}, \bibinfo{person}{Li Shu},
  \bibinfo{person}{Mudhakar Srivatsa}, \bibinfo{person}{Sofia Vallecorsa},
  {and} \bibinfo{person}{Kyongmin Yeo}.} \bibinfo{year}{2025}\natexlab{}.
\newblock \bibinfo{booktitle}{\emph{Fast Calorimeter Simulation Challenge 2022
  - Submissions Dataset 2}}.
\newblock
\href{https://doi.org/10.5281/zenodo.15962050}{doi:\nolinkurl{10.5281/zenodo.15962050}}


\bibitem[Favaro et~al\mbox{.}(2025)]%
        {favaro2025calodream}
\bibfield{author}{\bibinfo{person}{Luigi Favaro}, \bibinfo{person}{Ayodele
  Ore}, \bibinfo{person}{Sofia~Palacios Schweitzer}, {and}
  \bibinfo{person}{Tilman Plehn}.} \bibinfo{year}{2025}\natexlab{}.
\newblock \showarticletitle{{CaloDREAM}: Detector Response Emulation via
  Attentive Flow Matching}.
\newblock \bibinfo{journal}{\emph{SciPost Physics}}  \bibinfo{volume}{18}
  (\bibinfo{year}{2025}), \bibinfo{pages}{088}.
\newblock


\bibitem[Gao et~al\mbox{.}(2023)]%
        {gao2023prediff}
\bibfield{author}{\bibinfo{person}{Zhihan Gao}, \bibinfo{person}{Xingjian Shi},
  \bibinfo{person}{Boran Han}, \bibinfo{person}{Hao Wang},
  \bibinfo{person}{Xiaoyong Jin}, \bibinfo{person}{Danielle~C. Maddix},
  \bibinfo{person}{Yi Zhu}, \bibinfo{person}{Mu Li}, {and}
  \bibinfo{person}{Yuyang Wang}.} \bibinfo{year}{2023}\natexlab{}.
\newblock \showarticletitle{{PreDiff}: Precipitation Nowcasting with Latent
  Diffusion Models}. In \bibinfo{booktitle}{\emph{Advances in Neural
  Information Processing Systems (NeurIPS)}}.
\newblock


\bibitem[Hassan et~al\mbox{.}(2024)]%
        {hassan2024etflow}
\bibfield{author}{\bibinfo{person}{Majdi Hassan} {et~al\mbox{.}}}
  \bibinfo{year}{2024}\natexlab{}.
\newblock \showarticletitle{{ET-Flow}: Equivariant Flow-Matching for Molecular
  Conformer Generation}. In \bibinfo{booktitle}{\emph{Advances in Neural
  Information Processing Systems (NeurIPS)}}.
\newblock
\showeprint[arxiv]{2410.22388}


\bibitem[Ho et~al\mbox{.}(2020)]%
        {ho2020denoising}
\bibfield{author}{\bibinfo{person}{Jonathan Ho}, \bibinfo{person}{Ajay Jain},
  {and} \bibinfo{person}{Pieter Abbeel}.} \bibinfo{year}{2020}\natexlab{}.
\newblock \bibinfo{title}{Denoising Diffusion Probabilistic Models}.
\newblock
\showeprint[arxiv]{2006.11239}~[cs.LG]


\bibitem[Hoogeboom et~al\mbox{.}(2022)]%
        {hoogeboom2022edm}
\bibfield{author}{\bibinfo{person}{Emiel Hoogeboom},
  \bibinfo{person}{Victor~Garcia Satorras}, \bibinfo{person}{Cl{\'e}ment
  Vignac}, {and} \bibinfo{person}{Max Welling}.}
  \bibinfo{year}{2022}\natexlab{}.
\newblock \showarticletitle{Equivariant Diffusion for Molecule Generation in
  3D}. In \bibinfo{booktitle}{\emph{International Conference on Machine
  Learning (ICML)}}.
\newblock
\shownote{Oral presentation}.
\newblock


\bibitem[Jiang et~al\mbox{.}(2026)]%
        {jiang2026calotrilogy}
\bibfield{author}{\bibinfo{person}{Cheng Jiang}, \bibinfo{person}{Sitian Qian},
  \bibinfo{person}{Kevin Pedro}, \bibinfo{person}{Oz Amram},
  \bibinfo{person}{Huilin Qu}, {and} \bibinfo{person}{Maggie Voetberg}.}
  \bibinfo{year}{2026}\natexlab{}.
\newblock \showarticletitle{CaloTrilogy: Toward a Breakthrough in One-Step,
  End-to-End, Physics-Guided Shower Generation for Modern Calorimeters}.
\newblock \bibinfo{journal}{\emph{arXiv preprint arXiv:2606.04165}}
  (\bibinfo{year}{2026}).
\newblock


\bibitem[Kansal et~al\mbox{.}(2023)]%
        {kansal2023evaluating}
\bibfield{author}{\bibinfo{person}{Raghav Kansal}, \bibinfo{person}{Anni Li},
  \bibinfo{person}{Javier Duarte}, \bibinfo{person}{Nadezda Chernyavskaya},
  \bibinfo{person}{Maurizio Pierini}, \bibinfo{person}{Breno Orzari}, {and}
  \bibinfo{person}{Thiago Tomei}.} \bibinfo{year}{2023}\natexlab{}.
\newblock \showarticletitle{Evaluating generative models in high energy
  physics}.
\newblock \bibinfo{journal}{\emph{Physical Review D}} \bibinfo{volume}{107},
  \bibinfo{number}{7} (\bibinfo{year}{2023}), \bibinfo{pages}{076017}.
\newblock


\bibitem[Kendall et~al\mbox{.}(2018)]%
        {kendall2018multi}
\bibfield{author}{\bibinfo{person}{Alex Kendall}, \bibinfo{person}{Yarin Gal},
  {and} \bibinfo{person}{Roberto Cipolla}.} \bibinfo{year}{2018}\natexlab{}.
\newblock \showarticletitle{Multi-task learning using uncertainty to weigh
  losses for scene geometry and semantics}. In
  \bibinfo{booktitle}{\emph{Proceedings of the IEEE conference on computer
  vision and pattern recognition}}. \bibinfo{pages}{7482--7491}.
\newblock


\bibitem[Kobylianskii et~al\mbox{.}(2024)]%
        {kobylianskii2024calograph}
\bibfield{author}{\bibinfo{person}{Dmitrii Kobylianskii},
  \bibinfo{person}{Nathalie Soybelman}, \bibinfo{person}{Etienne Dreyer}, {and}
  \bibinfo{person}{Eilam Gross}.} \bibinfo{year}{2024}\natexlab{}.
\newblock \showarticletitle{{CaloGraph}: Graph-Based Diffusion Model for Fast
  Shower Generation in Calorimeters with Irregular Geometry}.
\newblock \bibinfo{journal}{\emph{Physical Review D}} (\bibinfo{year}{2024}).
\newblock
\showeprint[arxiv]{2402.11575}


\bibitem[Krause et~al\mbox{.}(2025)]%
        {krause2025calochallenge}
\bibfield{author}{\bibinfo{person}{Claudius Krause}, \bibinfo{person}{Michele
  Faucci~Giannelli}, \bibinfo{person}{Gregor Kasieczka},
  \bibinfo{person}{Benjamin Nachman}, \bibinfo{person}{Dalila Salamani},
  \bibinfo{person}{David Shih}, \bibinfo{person}{Anna Zaborowska},
  \bibinfo{person}{Oz Amram}, \bibinfo{person}{Kerstin Borras},
  \bibinfo{person}{Matthew~R Buckley}, {et~al\mbox{.}}}
  \bibinfo{year}{2025}\natexlab{}.
\newblock \showarticletitle{CaloChallenge 2022: a community challenge for fast
  calorimeter simulation}.
\newblock \bibinfo{journal}{\emph{Reports on Progress in Physics}}
  \bibinfo{volume}{88}, \bibinfo{number}{11} (\bibinfo{year}{2025}),
  \bibinfo{pages}{116201}.
\newblock


\bibitem[Krause and Shih(2021)]%
        {krause2021caloflow_2}
\bibfield{author}{\bibinfo{person}{Claudius Krause} {and}
  \bibinfo{person}{David Shih}.} \bibinfo{year}{2021}\natexlab{}.
\newblock \showarticletitle{CaloFlow II: Even faster and still accurate
  generation of calorimeter showers with normalizing flows}.
\newblock \bibinfo{journal}{\emph{arXiv preprint arXiv:2110.11377}}
  (\bibinfo{year}{2021}).
\newblock


\bibitem[Kynk{\"a}{\"a}nniemi et~al\mbox{.}(2019)]%
        {kynkaanniemi2019improved}
\bibfield{author}{\bibinfo{person}{Tuomas Kynk{\"a}{\"a}nniemi},
  \bibinfo{person}{Tero Karras}, \bibinfo{person}{Samuli Laine},
  \bibinfo{person}{Jaakko Lehtinen}, {and} \bibinfo{person}{Timo Aila}.}
  \bibinfo{year}{2019}\natexlab{}.
\newblock \showarticletitle{Improved precision and recall metric for assessing
  generative models}.
\newblock \bibinfo{journal}{\emph{Advances in neural information processing
  systems}}  \bibinfo{volume}{32} (\bibinfo{year}{2019}).
\newblock


\bibitem[Leigh et~al\mbox{.}(2024a)]%
        {leigh2024pcjedi}
\bibfield{author}{\bibinfo{person}{Matthew Leigh}, \bibinfo{person}{Debajyoti
  Sengupta}, \bibinfo{person}{Guillaume Qu{\'e}tant},
  \bibinfo{person}{John~Andrew Raine}, \bibinfo{person}{Knut Zoch}, {and}
  \bibinfo{person}{Tobias Golling}.} \bibinfo{year}{2024}\natexlab{a}.
\newblock \showarticletitle{{PC-JeDi}: Diffusion for Particle Cloud Generation
  in High Energy Physics}.
\newblock \bibinfo{journal}{\emph{SciPost Physics}}  \bibinfo{volume}{16}
  (\bibinfo{year}{2024}), \bibinfo{pages}{018}.
\newblock


\bibitem[Leigh et~al\mbox{.}(2024b)]%
        {leigh2024pcdroid}
\bibfield{author}{\bibinfo{person}{Matthew Leigh}, \bibinfo{person}{Debajyoti
  Sengupta}, \bibinfo{person}{John~Andrew Raine}, \bibinfo{person}{Guillaume
  Qu{\'e}tant}, {and} \bibinfo{person}{Tobias Golling}.}
  \bibinfo{year}{2024}\natexlab{b}.
\newblock \showarticletitle{{PC-Droid}: Faster Diffusion and Improved Quality
  for Particle Cloud Generation}.
\newblock \bibinfo{journal}{\emph{Physical Review D}} (\bibinfo{year}{2024}).
\newblock
\showeprint[arxiv]{2307.06836}


\bibitem[Li et~al\mbox{.}(2026)]%
        {li2026elign}
\bibfield{author}{\bibinfo{person}{Yunyang Li}, \bibinfo{person}{Lin Huang},
  \bibinfo{person}{Luojia Xia}, {et~al\mbox{.}}}
  \bibinfo{year}{2026}\natexlab{}.
\newblock \showarticletitle{Elign: Equivariant Diffusion Model Alignment from
  Foundational Machine Learning Force Fields}.
\newblock \bibinfo{journal}{\emph{arXiv preprint arXiv:2601.21985}}
  (\bibinfo{year}{2026}).
\newblock
\showeprint[arxiv]{2601.21985}~[cs.LG]


\bibitem[Liang et~al\mbox{.}(2025)]%
        {liang2025ccfm}
\bibfield{author}{\bibinfo{person}{Jinhao Liang}, \bibinfo{person}{Yixuan Sun},
  \bibinfo{person}{Anirban Samaddar}, \bibinfo{person}{Sandeep Madireddy},
  {and} \bibinfo{person}{Ferdinando Fioretto}.}
  \bibinfo{year}{2025}\natexlab{}.
\newblock \showarticletitle{Chance-Constrained Flow Matching}. In
  \bibinfo{booktitle}{\emph{Advances in Neural Information Processing Systems
  (NeurIPS)}}.
\newblock
\shownote{arXiv:2509.25157}.
\newblock


\bibitem[Liu et~al\mbox{.}(2021)]%
        {liu2021towards}
\bibfield{author}{\bibinfo{person}{Liyang Liu}, \bibinfo{person}{Yi Li},
  \bibinfo{person}{Zhanghui Kuang}, \bibinfo{person}{Jing-Hao Xue},
  \bibinfo{person}{Yimin Chen}, \bibinfo{person}{Wenming Yang},
  \bibinfo{person}{Qingmin Liao}, {and} \bibinfo{person}{Wayne Zhang}.}
  \bibinfo{year}{2021}\natexlab{}.
\newblock \showarticletitle{Towards impartial multi-task learning}. In
  \bibinfo{booktitle}{\emph{International conference on learning
  representations}}.
\newblock


\bibitem[Liu et~al\mbox{.}(2025)]%
        {liu2025config}
\bibfield{author}{\bibinfo{person}{Qiang Liu}, \bibinfo{person}{Mengyu Chu},
  {and} \bibinfo{person}{Nils Thuerey}.} \bibinfo{year}{2025}\natexlab{}.
\newblock \showarticletitle{ConFIG: Towards Conflict-free Training of Physics
  Informed Neural Networks}. In \bibinfo{booktitle}{\emph{International
  Conference on Learning Representations (ICLR)}}.
\newblock
\showeprint[arxiv]{2408.11104}


\bibitem[Mikuni and Nachman(2022)]%
        {mikuni2022score}
\bibfield{author}{\bibinfo{person}{Vinicius Mikuni} {and}
  \bibinfo{person}{Benjamin Nachman}.} \bibinfo{year}{2022}\natexlab{}.
\newblock \showarticletitle{Score-based generative models for calorimeter
  shower simulation}.
\newblock \bibinfo{journal}{\emph{Physical Review D}} \bibinfo{volume}{106},
  \bibinfo{number}{9} (\bibinfo{year}{2022}), \bibinfo{pages}{092009}.
\newblock


\bibitem[Mikuni and Nachman(2024)]%
        {mikuni2024caloscore2}
\bibfield{author}{\bibinfo{person}{Vinicius Mikuni} {and}
  \bibinfo{person}{Benjamin Nachman}.} \bibinfo{year}{2024}\natexlab{}.
\newblock \showarticletitle{CaloScore v2: single-shot calorimeter shower
  simulation with diffusion models}.
\newblock \bibinfo{journal}{\emph{Journal of Instrumentation}}
  \bibinfo{volume}{19}, \bibinfo{number}{02} (\bibinfo{year}{2024}),
  \bibinfo{pages}{P02001}.
\newblock
\href{https://doi.org/10.1088/1748-0221/19/02/P02001}{doi:\nolinkurl{10.1088/1748-0221/19/02/P02001}}


\bibitem[Naeem et~al\mbox{.}(2020)]%
        {naeem2020reliable}
\bibfield{author}{\bibinfo{person}{Muhammad~Ferjad Naeem},
  \bibinfo{person}{Seong~Joon Oh}, \bibinfo{person}{Youngjung Uh},
  \bibinfo{person}{Yunjey Choi}, {and} \bibinfo{person}{Jaejun Yoo}.}
  \bibinfo{year}{2020}\natexlab{}.
\newblock \showarticletitle{Reliable fidelity and diversity metrics for
  generative models}. In \bibinfo{booktitle}{\emph{International conference on
  machine learning}}. PMLR, \bibinfo{pages}{7176--7185}.
\newblock


\bibitem[Peebles and Xie(2023)]%
        {peebles2023scalable}
\bibfield{author}{\bibinfo{person}{William Peebles} {and}
  \bibinfo{person}{Saining Xie}.} \bibinfo{year}{2023}\natexlab{}.
\newblock \showarticletitle{Scalable diffusion models with transformers}. In
  \bibinfo{booktitle}{\emph{Proceedings of the IEEE/CVF international
  conference on computer vision}}. \bibinfo{pages}{4195--4205}.
\newblock


\bibitem[Platt and Barr(1988)]%
        {platt1988mdmm}
\bibfield{author}{\bibinfo{person}{John~C. Platt} {and}
  \bibinfo{person}{Alan~H. Barr}.} \bibinfo{year}{1988}\natexlab{}.
\newblock \showarticletitle{Constrained Differential Optimization}. In
  \bibinfo{booktitle}{\emph{Neural Information Processing Systems (NIPS)}},
  \bibfield{editor}{\bibinfo{person}{Dana~Z. Anderson}} (Ed.),
  Vol.~\bibinfo{volume}{1}. \bibinfo{publisher}{American Institute of Physics},
  \bibinfo{pages}{612--621}.
\newblock


\bibitem[Raikwar et~al\mbox{.}(2024)]%
        {raikwar2024calodit}
\bibfield{author}{\bibinfo{person}{Piyush Raikwar}, \bibinfo{person}{Renato
  Cardoso}, \bibinfo{person}{Kristina Jaruskova}, \bibinfo{person}{Dalila
  Salamani}, \bibinfo{person}{Sofia Vallecorsa}, \bibinfo{person}{Anna
  Zaborowska}, {et~al\mbox{.}}} \bibinfo{year}{2024}\natexlab{}.
\newblock \bibinfo{title}{{CaloDiT}: Diffusion with transformers for fast
  shower simulation}.
\newblock \bibinfo{howpublished}{Presentation at the 22nd International
  Workshop on Advanced Computing and Analysis Techniques in Physics Research
  (ACAT)}.
\newblock
\shownote{\url{https://indico.cern.ch/event/1330797/contributions/5796591/}}.
\newblock


\bibitem[Raikwar et~al\mbox{.}(2025)]%
        {raikwar2025generalisable}
\bibfield{author}{\bibinfo{person}{Piyush Raikwar}, \bibinfo{person}{Anna
  Zaborowska}, \bibinfo{person}{Peter McKeown}, \bibinfo{person}{Renato
  Cardoso}, \bibinfo{person}{Mikolaj Piorczynski}, {and}
  \bibinfo{person}{Kyongmin Yeo}.} \bibinfo{year}{2025}\natexlab{}.
\newblock \showarticletitle{A Generalisable Generative Model for Multi-Detector
  Calorimeter Simulation}.
\newblock \bibinfo{journal}{\emph{arXiv preprint arXiv:2509.07700}}
  (\bibinfo{year}{2025}).
\newblock


\bibitem[Shuman et~al\mbox{.}(2013)]%
        {shuman2013emerging}
\bibfield{author}{\bibinfo{person}{David~I Shuman}, \bibinfo{person}{Sunil~K
  Narang}, \bibinfo{person}{Pascal Frossard}, \bibinfo{person}{Antonio Ortega},
  {and} \bibinfo{person}{Pierre Vandergheynst}.}
  \bibinfo{year}{2013}\natexlab{}.
\newblock \showarticletitle{The emerging field of signal processing on graphs:
  Extending high-dimensional data analysis to networks and other irregular
  domains}.
\newblock \bibinfo{journal}{\emph{IEEE signal processing magazine}}
  \bibinfo{volume}{30}, \bibinfo{number}{3} (\bibinfo{year}{2013}),
  \bibinfo{pages}{83--98}.
\newblock


\bibitem[Song et~al\mbox{.}(2021)]%
        {song2021scorebased}
\bibfield{author}{\bibinfo{person}{Yang Song}, \bibinfo{person}{Jascha
  Sohl-Dickstein}, \bibinfo{person}{Diederik~P. Kingma},
  \bibinfo{person}{Abhishek Kumar}, \bibinfo{person}{Stefano Ermon}, {and}
  \bibinfo{person}{Ben Poole}.} \bibinfo{year}{2021}\natexlab{}.
\newblock \bibinfo{title}{Score-Based Generative Modeling through Stochastic
  Differential Equations}.
\newblock
\showeprint[arxiv]{2011.13456}~[cs.LG]


\bibitem[Tian et~al\mbox{.}(2025)]%
        {tian2025equiflow}
\bibfield{author}{\bibinfo{person}{Qingwen Tian}, \bibinfo{person}{Yuxin Xu},
  \bibinfo{person}{Yixuan Yang}, \bibinfo{person}{Zhen Wang},
  \bibinfo{person}{Ziqi Liu}, \bibinfo{person}{Pengju Yan}, {and}
  \bibinfo{person}{Xiaolin Li}.} \bibinfo{year}{2025}\natexlab{}.
\newblock \showarticletitle{{EquiFlow}: Equivariant Conditional Flow Matching
  with Optimal Transport for 3D Molecular Conformation Prediction}. In
  \bibinfo{booktitle}{\emph{AAAI Conference on Artificial Intelligence}}.
\newblock


\bibitem[Trupin et~al\mbox{.}(2026)]%
        {trupin2026learning}
\bibfield{author}{\bibinfo{person}{Noah Trupin}, \bibinfo{person}{Rahul Ghosh},
  {and} \bibinfo{person}{Aadi Jangid}.} \bibinfo{year}{2026}\natexlab{}.
\newblock \showarticletitle{Learning Flow Distributions via
  Projection-Constrained Diffusion on Manifolds}.
\newblock \bibinfo{journal}{\emph{arXiv preprint arXiv:2602.17773}}
  (\bibinfo{year}{2026}).
\newblock


\bibitem[Utkarsh et~al\mbox{.}(2026)]%
        {utkarsh2026physics}
\bibfield{author}{\bibinfo{person}{Utkarsh Utkarsh}, \bibinfo{person}{Pengfei
  Cai}, \bibinfo{person}{Alan Edelman}, \bibinfo{person}{Rafael
  Gomez-Bombarelli}, {and} \bibinfo{person}{Christopher Rackauckas}.}
  \bibinfo{year}{2026}\natexlab{}.
\newblock \showarticletitle{Physics-constrained flow matching: Sampling
  generative models with hard constraints}.
\newblock \bibinfo{journal}{\emph{Advances in Neural Information Processing
  Systems}}  \bibinfo{volume}{38} (\bibinfo{year}{2026}),
  \bibinfo{pages}{160217--160252}.
\newblock


\bibitem[Vaitl and Klein(2025)]%
        {vaitl2025pathgrad}
\bibfield{author}{\bibinfo{person}{Lorenz Vaitl} {and} \bibinfo{person}{Leon
  Klein}.} \bibinfo{year}{2025}\natexlab{}.
\newblock \showarticletitle{Path Gradients after Flow Matching}. In
  \bibinfo{booktitle}{\emph{Advances in Neural Information Processing Systems
  (NeurIPS)}}.
\newblock
\showeprint[arxiv]{2505.10139}~[cs.LG]


\bibitem[Wang et~al\mbox{.}(2021)]%
        {wang2021pathologies}
\bibfield{author}{\bibinfo{person}{Sifan Wang}, \bibinfo{person}{Yujun Teng},
  {and} \bibinfo{person}{Paris Perdikaris}.} \bibinfo{year}{2021}\natexlab{}.
\newblock \showarticletitle{Understanding and Mitigating Gradient Flow
  Pathologies in Physics-Informed Neural Networks}.
\newblock \bibinfo{journal}{\emph{SIAM Journal on Scientific Computing}}
  \bibinfo{volume}{43}, \bibinfo{number}{5} (\bibinfo{year}{2021}),
  \bibinfo{pages}{A3055--A3081}.
\newblock


\bibitem[Yan et~al\mbox{.}(2023)]%
        {yan2023auxiliary}
\bibfield{author}{\bibinfo{person}{Junjun Yan}, \bibinfo{person}{Xinhai Chen},
  \bibinfo{person}{Zhichao Wang}, \bibinfo{person}{Enqiang Zhou}, {and}
  \bibinfo{person}{Jie Liu}.} \bibinfo{year}{2023}\natexlab{}.
\newblock \showarticletitle{Auxiliary-tasks learning for physics-informed
  neural network-based partial differential equations solving}.
\newblock \bibinfo{journal}{\emph{arXiv preprint arXiv:2307.06167}}
  (\bibinfo{year}{2023}).
\newblock


\bibitem[Yu et~al\mbox{.}(2020)]%
        {yu2020pcgrad}
\bibfield{author}{\bibinfo{person}{Tianhe Yu}, \bibinfo{person}{Saurabh Kumar},
  \bibinfo{person}{Abhishek Gupta}, \bibinfo{person}{Sergey Levine},
  \bibinfo{person}{Karol Hausman}, {and} \bibinfo{person}{Chelsea Finn}.}
  \bibinfo{year}{2020}\natexlab{}.
\newblock \showarticletitle{Gradient Surgery for Multi-Task Learning}. In
  \bibinfo{booktitle}{\emph{Advances in Neural Information Processing Systems
  (NeurIPS)}}, Vol.~\bibinfo{volume}{33}. \bibinfo{pages}{5824--5836}.
\newblock


\bibitem[Zhang and Zou(2025)]%
        {zhang2025physics}
\bibfield{author}{\bibinfo{person}{Yi Zhang} {and} \bibinfo{person}{Difan
  Zou}.} \bibinfo{year}{2025}\natexlab{}.
\newblock \showarticletitle{Physics-informed distillation of diffusion models
  for pde-constrained generation}.
\newblock \bibinfo{journal}{\emph{arXiv preprint arXiv:2505.22391}}
  (\bibinfo{year}{2025}).
\newblock


\end{thebibliography}

\appendix

\section{Evaluation Protocol and Metric Details}
\label{app:eval}

\subsection{Samples and Reference Data}
\label{app:reference}

All metrics are computed against the second file of CaloChallenge
Dataset~2, which contains 100{,}000 Geant4 electron showers held out
from training~\cite{krause2025calochallenge}. The same reference
file is used for every model, including the prior surrogates we
re-evaluate for CFD. For each of our trained models, we generate
100{,}000 showers with incident energies drawn from the reference
distribution, matching the reference sample size.

For the prior surrogates, the standard metrics---FPD, KPD, PRDC,
and the classifier test---are taken directly from the CaloChallenge
report~\cite{krause2025calochallenge} (Tables~C15, C16, and~C19),
which evaluated all submissions with the official pipeline against
this same reference file. CFD is not reported there, so we compute
it ourselves from the public submission samples released on
Zenodo~\cite{faucci_giannelli_2025_15962050}, using the identical procedure
applied to our own models (Appendix~\ref{app:cfd-impl}). 

CaloDiT-2 postdates the CaloChallenge evaluation; its released artifacts include pre-trained models but no generated sample sets, so its CFD is omitted and we leave regeneration from the Geant4-distributed checkpoints to future work.

\subsection{Seeds and Aggregation}
\label{app:seeds}
The three-seed protocol and the fixed energy network are described in
Section~\ref{sec:setup}; the reported spread therefore reflects shape-network
initialization alone. Every metric for a trained model is reported as
mean~$\pm$~std over the three seeds, each evaluated on its own generated sample
set. Prior-surrogate rows are single values from the CaloChallenge report,
carrying that report's uncertainties where available.

\subsection{Standard Metric Pipeline}
\label{app:pipeline}

We compute FPD, KPD, and the classifier test with the official
CaloChallenge evaluation pipeline
(\texttt{evaluate.py})~\cite{calochallenge_homepage}, run unmodified,
so that our trained models and the prior surrogates are scored
under one protocol. Metric definitions follow the CaloChallenge
report~\cite{krause2025calochallenge}. FPD and KPD are computed on
the standard set of physics-motivated shower features and reported
scaled by $10^3$, as in the main text.

\subsection{Classifier Test Protocol}
\label{app:classifier}

The classifier test trains a binary classifier to separate generated
showers (label 0) from Geant4 reference showers (label 1) and
reports how well it succeeds. We evaluate two feature sets. The
low-level classifier uses the voxel energies, each shower normalized
by its incident energy. The high-level classifier uses the
shower-shape observables of the CaloChallenge
protocol~\cite{krause2025calochallenge}: total deposited energy,
per-layer energies, and the energy-weighted centroids and widths in
the two angular directions, together with the incident energy. Both
feature sets apply the low-energy cut of 0.015~MeV used throughout
our evaluation.

The classifier is the standard CaloChallenge
network~\cite{krause2025calochallenge}, a multilayer perceptron
with two hidden layers of 512 units, trained with the pipeline's
default optimizer, learning rate, and batch size.
The combined samples are split into equal-sized classes and then
into training, test, and validation partitions. We select the
checkpoint with the best test accuracy and report the area under
the ROC curve (AUC) on the held-out validation partition; an AUC
near 0.5 indicates the two sample sets are indistinguishable. We
report AUC averaged over three independent runs, each using samples
from a separately seeded shape network, so the spread reflects
variation across trained models. Prior-surrogate AUC values are
taken from the CaloChallenge report, which averages ten evaluation
runs~\cite{krause2025calochallenge}.

\subsection{PRDC Settings}
\label{app:prdc}

Precision, recall, density, and coverage
(PRDC)~\cite{naeem2020reliable} are computed with $k = 5$ nearest
neighbors, matching the CaloChallenge protocol, on the same feature
space as the high-level classifier. Precision and density measure
sample quality (generated samples lying on the reference manifold);
recall and coverage measure diversity (reference samples reachable
from the generated manifold). Prior-surrogate values are taken from
Table~C19 of the CaloChallenge report~\cite{krause2025calochallenge};
trained models report mean~$\pm$~std over three seeds. Results are
given in Table~\ref{tab:prdc} in Appendix~\ref{app:prdc-schedule}.

\subsection{CFD Implementation Details}
\label{app:cfd-impl}
CFD is computed with a single procedure for every model in
Tables~\ref{tab:main} and~\ref{tab:prdc}: from the public Zenodo submission
samples for the prior surrogates, and from samples drawn from our own trained
models otherwise. Correlations are computed on the voxel energies stored in the
sample files (physical units, MeV), which already incorporate the 0.015~MeV
low-energy cut; no further thresholding is applied at the correlation stage. For
each model the generated and reference sets are truncated to a common size of
100{,}000 showers. A transverse position enters the voxel-wise computation only
if it carries nonzero variance in both the reference and generated samples,
since the Pearson correlation is otherwise undefined. For Dataset~2 with
100{,}000 showers every position satisfies this, so all
$44 \times 16 \times 9 = 6{,}336$ layer-pair positions are retained and none are
discarded, and both Frobenius sums in Eq.~\eqref{eq:cfd-vox} run over this full
set.
\subsection{Layer-wise CFD}
\label{app:cfd-layer}
Layer-wise CFD measures the global correlation structure between per-layer
energies. For each pair of layers $(i, j)$ with $i, j = 1, \dots, L$, we compute
the Pearson correlation across showers between the layer energies at $i$ and $j$:
\begin{equation}
  \rho_{ij}
  = \frac{\sum_k \big(E_{i,k} - \bar{E}_i\big)\big(E_{j,k} - \bar{E}_j\big)}
         {\sqrt{\sum_k \big(E_{i,k} - \bar{E}_i\big)^2}\;
          \sqrt{\sum_k \big(E_{j,k} - \bar{E}_j\big)^2}} ,
  \label{eq:cfd-layer-corr}
\end{equation}
where $E_{i,k}$ is the energy deposited in layer $i$ by shower $k$, the index $k$
runs over the $K$ showers in the set, and
$\bar{E}_i = \tfrac{1}{K}\sum_k E_{i,k}$ is the mean layer energy across showers.
Collecting $\rho_{ij}$ over all layer pairs gives an $L \times L$ correlation
matrix $\rho^\text{ref}$ for \textsc{Geant4} and $\rho^\text{gen}$ for the
generated samples. As in the voxel-wise case, we evaluate only pairs where both
layers carry nonzero variance, since the correlation is undefined otherwise.
Layer-wise CFD is the normalized Frobenius distance between the two matrices,
restricted to the off-diagonal entries,
\begin{equation}
  \mathrm{CFD}_\text{layer}
  = \frac{\big\| \rho^\text{gen} - \rho^\text{ref} \big\|_{F,\,\text{off}}}
         {\big\| \rho^\text{ref} \big\|_{F,\,\text{off}}} ,
  \qquad
  \big\| \rho \big\|_{F,\,\text{off}}
  = \sqrt{\sum_{i \neq j} \big(\rho_{ij}\big)^2}\, ,
  \label{eq:cfd-layer}
\end{equation}
where both sums run over $i \neq j$. The unit diagonal is excluded because it is
identical in both matrices and would only dilute the normalized distance. Lower
values indicate closer agreement with \textsc{Geant4}'s inter-layer correlation
structure. The voxel-wise form (Eq.~\eqref{eq:cfd-vox}, main text) applies the
same construction to the per-position correlations between consecutive layers.


%
%

\section{Data Representation and Model Architecture}
\label{app:arch}

\subsection{Data Representation and Preprocessing}
\label{app:preproc}

A shower is stored as physical voxel energies $x\in\mathbb{R}^{V}_{\ge 0}$ in MeV,
and the networks act on the preprocessed representation $\tilde{x}=\mathcal{T}(x)$
defined in Section~\ref{sec:formulation}. We reuse the transform of
CaloDREAM~\cite{favaro2025calodream}, built from three invertible maps. Every
voxel is first divided by its layer energy, so the shape network sees only the
within-layer shares $s\in[0,1]$ and the layer energies are handled separately by
the ratio vector $u$. Each share then passes through a regularized logit,
\begin{equation}
s' = \log\frac{s_\alpha}{1-s_\alpha},\qquad
s_\alpha=(1-2\alpha)\,s+\alpha,\qquad \alpha=10^{-6},
\label{eq:preproc}
\end{equation}
which pushes the many near-zero voxels onto a usable range while staying
one-to-one, and each feature is centered and scaled using fixed training-set statistics. Generation runs $\mathcal{T}^{-1}$ and then rescales every layer's
voxels to the energy set by $u$, fixing the total and per-layer energies of each
generated shower exactly.

The ratio vector $u\in\mathbb{R}^{L}$ expresses the layer energies
$E_0,\dots,E_{L-1}$ through $E_{\mathrm{inc}}$, following the parameterization of
CaloFlow~\cite{krause2021caloflow_2} adopted by
CaloDREAM~\cite{favaro2025calodream},
\begin{equation}
u_0=\frac{\sum_i E_i}{f\,E_{\mathrm{inc}}},\qquad
u_i=\frac{E_i}{\sum_{j\ge i}E_j},\quad i=1,\dots,L-1,
\label{eq:u-def}
\end{equation}
where the constant $f=2.85$ keeps $u_0\in[0,1]$. For $i>0$, $u_i$ is the share of
the energy still undeposited at layer $i$ that layer $i$ receives; the map
$(E_{\mathrm{inc}},u)\mapsto(E_0,\dots,E_{L-1})$ therefore has a closed-form
inverse and returns the layer energies without approximation.

\begin{table*}[t]
\centering
\small
\caption{Hyperparameters of the energy and shape networks on Dataset~2.}
\label{tab:hyperparams}
\begin{tabular}[t]{@{}ll@{}}
\multicolumn{2}{c}{\textbf{Energy network}}\\
\hline
\textbf{Parameter} & \textbf{DS2} \\
\hline
Epochs & 500 \\
LR sched. & one-cycle \\
Max LR & $2\times 10^{-3}$ \\
Batch size & 4000 \\
Noise schedule & Linear \\
Solver & DDPM ancestral sampler \\
\hline
\multicolumn{2}{@{}l}{\textit{Conditioning network (Transformer)}} \\
\hline
Dim embedding & 64 \\
Intermediate dim & 512 \\
Num heads & 4 \\
Encoder layers & 4 \\
Decoder layers & 4 \\
Normalization & LayerNorm \\
\hline
\multicolumn{2}{@{}l}{\textit{Denoising network (dense feed-forward)}} \\
\hline
Intermediate dim & 256 \\
Num layers & 8 \\
Activation & SiLU \\
\hline
\multicolumn{2}{@{}l}{\textit{Training}} \\
\hline
Optimizer & AdamW \\
Betas & (0.9, 0.999) \\
Timesteps & 1000 \\
\hline
\end{tabular}
\hspace{0.06\linewidth}
\begin{tabular}[t]{@{}ll@{}}
\multicolumn{2}{c}{\textbf{Shape network}}\\
\hline
\textbf{Parameter} & \textbf{DS2} \\
\hline
Epochs & 800 \\
LR sched. & CosineAnnealing \\
LR / Max LR & $5\times 10^{-4}$ / $1\times 10^{-3}$ \\
Batch size & 64 \\
Noise schedule & sigmoid \\
Solver & DDPM ancestral sampler \\
\hline
\multicolumn{2}{@{}l}{\textit{Backbone network (ViT)}} \\
\hline
Input shape & [1, 45, 16, 9] \\
Patch shape & [3, 16, 1] \\
Hidden dim & 240 \\
Depth (layers) & 6 \\
Num heads & 6 \\
MLP ratio & 4.0 \\
Condition dim & 46 \\
Attention & Standard (non-causal) \\
Prediction type & noise \\
\hline
\multicolumn{2}{@{}l}{\textit{Training}} \\
\hline
Optimizer & AdamW \\
Betas & (0.9, 0.999) \\
Epsilon & $1\times 10^{-6}$ \\
Weight decay & $1\times 10^{-5}$ \\
Timesteps & 1000 \\
\hline
\end{tabular}
\end{table*}

\subsection{Diffusion Background}
\label{app:ddpm}

Both networks are denoising diffusion probabilistic models~\cite{ho2020denoising}.
The forward process noises a clean sample $z_0$ into $z_t$ as in
Eq.~\eqref{eq:forward} of the main text, with $\alpha_t=1-\beta_t$ and
$\bar{\alpha}_t=\prod_{s\le t}\alpha_s$ fixed by the variance schedule
$\{\beta_t\}_{t=1}^{T}$. Sampling inverts this process one step at a time: a
noise-prediction network $\epsilon_\theta(z_t,t,c)$, conditioned on the timestep
$t$ and context $c$ (the incident energy, and for the shape network also $u$),
drives the ancestral update
\begin{equation}
z_{t-1}=\frac{1}{\sqrt{\alpha_t}}\!\left(z_t-\frac{1-\alpha_t}
{\sqrt{1-\bar{\alpha}_t}}\,\epsilon_\theta(z_t,t,c)\right)+\sigma_t\,\xi,
\qquad
\sigma_t^{2}=\beta_t\,\frac{1-\bar{\alpha}_{t-1}}{1-\bar{\alpha}_t},
\label{eq:ddpm-reverse}
\end{equation}
with $\xi\sim\mathcal{N}(0,I)$ for $t>1$ and $\xi=0$ at $t=1$ for a deterministic
final step. Both networks are trained with the standard noise-prediction
objective
\begin{equation}
\mathcal{L}_{\mathrm{denoise}}=\mathbb{E}_{z_0,\epsilon,t}
\big[\,\|\epsilon-\epsilon_\theta(z_t,t,c)\|_2^2\,\big].
\label{eq:ddpm-loss}
\end{equation}

\subsection{Energy Network: Autoregressive Transformer}
\label{app:energy}

The energy network predicts the layer-energy ratios $u=(u_0,\dots,u_{L-1})$
autoregressively, reflecting that the energy reaching a layer is constrained by
what has already been deposited upstream. Following the transfusion design of
CaloDREAM~\cite{favaro2025calodream}, it couples a transformer encoder for the
incident-energy condition with a causally masked decoder over layers and a
per-layer denoising head.

\paragraph{Conditioning and layer inputs.}
The incident energy $E_{\mathrm{inc}}$ is embedded as a single condition token and
so the encoder's self-attention runs over one element and reduces to the identity. Each layer $i$ is marked by a positional token $p_i$ that encodes its layer index, zero-padded to the token width; the decoder applies a causal mask, so layer $i$ attends only to the tokens
$\{p_0, \dots, p_{i-1}\}$ that precede it. A cross-attention between
the encoded condition and the decoder states produces a per-layer context $c_i$,
\begin{equation}
c_i=
\begin{cases}
f(E_{\mathrm{inc}}) & i=0,\\[2pt]
f(\mathbf{p}_0,\dots,\mathbf{p}_{i-1},E_{\mathrm{inc}}) & i>0,
\end{cases}
\label{eq:energy-context}
\end{equation}
that carries what is needed to predict the ratio at layer $i$.

\paragraph{Per-layer denoising.}
A single denoising head $\epsilon_\phi$ acts on the noised ratio $u_i(t)$ of each
layer, conditioned on its context $c_i$ and the timestep $t$, and is applied
across all $L$ layers,
\begin{equation}
\hat{\epsilon}\big(u(t),t,E_{\mathrm{inc}}\big)=
\big(\epsilon_\phi(u_0(t),c_0,t),\,\dots,\,\epsilon_\phi(u_{L-1}(t),c_{L-1},t)\big).
\label{eq:energy-eps}
\end{equation}
It is trained with the denoising objective of Eq.~\eqref{eq:ddpm-loss}, taking
$z_0=u$. Because the clean ratios are known for every layer, training evaluates
all layers in parallel; sampling instead proceeds layer by layer, each predicted
ratio conditioning the next.

\subsection{Shape Network: Conditional ViT}
\label{app:shape}

The shape network is a three-dimensional vision
transformer~\cite{peebles2023scalable}, adapted to calorimeter showers as in
CaloDREAM~\cite{favaro2025calodream}, that denoises the preprocessed shower
conditioned on $(E_{\mathrm{inc}},u)$. The calorimeter volume is split into
non-overlapping voxel patches; each patch is linearly embedded, tagged with a
learnable positional encoding, and processed by a stack of transformer blocks of
multi-head self-attention and feed-forward layers. A final linear layer maps each
patch back to voxel space as a noise estimate, and the patches are reassembled
into the shower volume. It is trained with the objective of
Eq.~\eqref{eq:ddpm-loss}, taking $z_0=\tilde{x}$; the input and patch dimensions
are listed in Table~\ref{tab:hyperparams}.

The conditioning inputs, the timestep $t$, the incident energy $E_{\mathrm{inc}}$,
and the ratio vector $u$, are embedded by small dense networks and summed into a
single vector that modulates every block. The modulation is affine: the condition
sets scale and shift parameters $(a,b)$ and a residual gate $\gamma$ applied to
both sublayers,
\begin{equation}
x_h = x + \gamma_h\, g_h(a_h x + b_h),\qquad
x_l = x_h + \gamma_l\, g_l(a_l x_h + b_l),
\label{eq:vit-block}
\end{equation}
where $g_h$ and $g_l$ are the self-attention and feed-forward operations and
$(a,b,\gamma)$ depend on the conditioning vector~\cite{peebles2023scalable}.
Patch size trades spatial detail against attention cost; the value used for
Dataset~2 is given in Table~\ref{tab:hyperparams}.

\subsection{Training and Hyperparameters}
\label{app:training}

As noted earlier, the Autoregressive Layer Network and the Shape Network are
trained independently but coupled during generation. All experiments were
conducted on a single NVIDIA Tesla V100-SXM2 GPU with 32\,GB of memory, using
CUDA~12.8 for both training and inference, with all models implemented in
PyTorch. Table~\ref{tab:hyperparams} reports the hyperparameters for both the
energy and shape networks.

%
\section{Auxiliary-Loss Combination Baselines}
\subsection{Gradient-Combination Baselines}
\label{app:gradbaselines}

The combination baselines isolate the rule that merges the denoising and
auxiliary gradients. Architecture, auxiliary losses, warm-up/decay schedule,
preprocessing, and the AdamW optimizer are all held fixed at the \textsc{Base}
configuration; only the combination rule changes, so any difference between these
rows reflects the rule alone. At every optimizer step we backpropagate the
denoising loss and each active physics loss \emph{separately} to full-parameter
gradient vectors. We index the objectives by $i=1,\dots,m$, writing
$\mathcal{L}_i$ for the $i$-th loss and $g_i=\nabla_\theta\mathcal{L}_i$ for its
gradient over the shape-network parameters $\theta$; the denoising objective is
fixed at $i=1$, so $g_1=g_d$ and the remaining $g_i$ are the physics-loss
gradients ($g_a$ in the two-objective case). Each rule below maps the set
$\{g_i\}$ to a single update that is written into the parameters before the AdamW
step. Unlike the weighted-sum baseline, these methods combine the per-objective
gradients directly, so the fixed per-loss weights $\lambda_i$ play no role (GradNorm
instead learns its own weights, described below). If any per-objective gradient is non-finite, we fall back to the
pure denoising gradient $g_d$ for that step. In our experiments two objectives
participate: the denoising loss and one physics loss (voxel or Laplacian).

\paragraph{ConFIG.}
We apply the conflict-free update of~\cite{liu2025config} through the authors'
\texttt{conflictfree} library. Each step backpropagates the denoising loss and the
active physics loss separately to full-parameter gradients $g_d$ and $g_a$, which
are passed to \texttt{ConFIG\_update} (pseudo-inverse solver,
$\texttt{rtol}=10^{-4}$, default equal-weight and projection-length models). The
returned direction has equal, positive projection onto both gradients---the unit
bisector of $g_d$ and $g_a$ in our two-objective case---and is written into the
parameters before the usual AdamW step. If either gradient is non-finite we fall
back to the pure denoising gradient. Architecture, auxiliary loss, schedule,
preprocessing, and optimizer are identical to \textsc{Base}, so this configuration
isolates the combination rule.

\paragraph{PCGrad.}
We apply gradient surgery~\cite{yu2020pcgrad} through the
conflictfree library. Given the per-objective gradients, any pair with a
negative inner product is deconflicted by projecting one gradient onto the normal
plane of the other; the deconflicted gradients are then summed and written into
the parameters. We log the number of conflicting pairs resolved per step. As for
the other rules, a non-finite gradient triggers the pure-denoising fallback.

\paragraph{IMTL-G.}
We use the impartial multi-task gradient rule~\cite{liu2021towards} through the
\texttt{IMTLGOperator} of the \texttt{conflictfree} library. It selects the
combined direction whose projection onto every objective's unit gradient
$\hat{u}_i=g_i/\lVert g_i\rVert$ is equal, so no single objective dominates the
shared update. We verify this impartiality online by logging the projection
$\hat{u}_i\cdot g_{\mathrm{c}}$ of the combined update $g_{\mathrm{c}}$ onto each
objective, which the rule drives to a common value. Non-finite gradients trigger the pure-denoising
fallback.

\paragraph{GradNorm.}
Unlike the three conflict-based rules, GradNorm~\cite{chen2018gradnorm} rebalances
gradient \emph{magnitudes} rather than resolving directional conflict. Each
objective receives a weight $w_i$ formed as a softmax over learned logits,
normalized to sum to the number of objectives and floored at $0.5$, and the
applied update is the weighted sum $\sum_i w_i\,g_i$ with the weights detached---equivalent to
backpropagating $\sum_i w_i\,\mathcal{L}_i$. The logits are trained (Adam,
learning rate $10^{-3}$) so that each objective's gradient norm is pulled toward
the batch-mean norm rescaled by its relative inverse training rate raised to the
power $\alpha=1.5$; the training rate is measured against a per-objective baseline $L_{0,i}$
(the mean of $\mathcal{L}_i$ over the first $100$ batches), during which an
equal-weight sum is used.
The gradient norms driving this weight update are measured over the full set of
shape-network parameters, rather than at a single last shared layer as in the
original formulation. GradNorm therefore behaves like the weighted-sum baseline but with weights adapted
online in place of the fixed $\lambda_i$. A non-finite gradient triggers the
pure-denoising fallback.

\subsection{Loss-weighting baselines}
These methods reduce the objectives to one scalar loss and backpropagate it once. They were our primary investigation before we turned to gradient-level
combination: because a scalar weight sets only the \emph{magnitude} of each loss, they cannot reorient the physics gradient away from a direction that conflicts with denoising---the limitation that motivated the gradient methods above.

\paragraph{Weighted sum.}
The simplest baseline forms a fixed linear combination
$\mathcal{L}=\sum_i \lambda_i \mathcal{L}_i$ and backpropagates it once. We set
$\lambda_1=1$ for the denoising loss and a small fixed weight for the physics loss
($\lambda_{\mathrm{lap}}=10^{-2}$ for the Laplacian loss and
$\lambda_{\mathrm{vox}}=10^{-4}$ for the voxel loss). These weights are constant
throughout training and, unlike the other methods, are not gated by the
warm-up/decay schedule. A single scalar per loss fixes both the magnitude and the
direction of the auxiliary contribution, so the physics gradient is applied in
full even where it opposes denoising.

\paragraph{Uncertainty weighting.}
Following the homoscedastic formulation of Kendall et al.~\cite{kendall2018multi}, we replace
the fixed weights with learned per-loss uncertainties. With a learnable
log-variance $s_i=\log\sigma_i^2$ per objective, each loss enters as
$\tfrac{1}{2}e^{-s_i}\mathcal{L}_i+\tfrac{1}{2}s_i$: the precision $e^{-s_i}$
scales the loss while the $\tfrac{1}{2}s_i$ term regularizes the variance. The
$s_i$ are initialized to zero and trained jointly with the network by the same
AdamW optimizer. We evaluate two variants. The \emph{pure} form applies the
learned weighting throughout training, as in the original method; the
\emph{scheduled} form additionally places the auxiliary term under the same
warm-up/decay envelope used by the other methods, so the schedule gates
\emph{when} the physics loss participates while its relative weight remains
\emph{learned}. In both, uncertainty weighting still reduces the two objectives to
a single scalar loss, so it rescales the physics gradient but cannot reorient it.
\section{Additional Results}
\subsection{PRDC}
\label{app:prdc-schedule}
\begin{table*}[t]
\centering
\caption{Precision, density, recall, and coverage (PRDC) on CaloChallenge
Dataset 2, computed with $k=5$ nearest neighbors. Prior-surrogate values are
taken from the CaloChallenge report \cite{krause2025calochallenge}. Trained
models report mean $\pm$ std over three seeds.}
\label{tab:prdc}
\begin{tabular}{llcccc}
\toprule
Model & Loss & Precision $\uparrow$ & Density$\uparrow$ & Recall $\uparrow$ & Coverage $\uparrow$ \\
\midrule
CaloDiffusion \cite{amram2023calodiffusion} & N/A & $0.239$ & $1.236$ & $0.235$ & $0.933$ \\
CaloScore~v2 \cite{mikuni2024caloscore2}    & N/A & $0.228$ & $1.013$ & $0.228$ & $0.933$ \\
CaloDiT \cite{raikwar2024calodit}           & N/A & $0.500$ & $10.23$ & $0.060$ & $0.924$ \\
CaloDREAM \cite{favaro2025calodream}        & N/A & $0.253$ & $1.146$ & $0.220$ & $0.976$ \\
\midrule
\base                & None      & $0.264 \pm 0.003$ & $1.330 \pm 0.040$ & $0.210 \pm 0.002$ & $0.984 \pm 0.002$ \\
\midrule
\basex{PCGrad}       & Voxel     & $0.273 \pm 0.013$ & $1.592 \pm 0.339$ & $0.199 \pm 0.016$ & $0.986 \pm 0.004$ \\
\basex{GradNorm}     & Voxel     & $0.280 \pm 0.001$ & $1.874 \pm 0.073$ & $0.178 \pm 0.002$ & $0.994 \pm 0.001$ \\
\basex{IMTL-G}       & Voxel     & $0.326 \pm 0.015$ & $5.625 \pm 0.741$ & $0.178 \pm 0.021$ & $0.891 \pm 0.032$ \\
\basex{ConFIG}       & Voxel     & $0.304 \pm 0.001$ & $2.674 \pm 0.174$ & $0.155 \pm 0.010$ & $0.994 \pm 0.001$ \\
\lantern             & Voxel     & $0.256 \pm 0.001$ & $1.309 \pm 0.052$ & $0.216 \pm 0.004$ & $0.981 \pm 0.002$ \\
\midrule
\basex{PCGrad}       & Laplacian & $0.293 \pm 0.017$ & $2.230 \pm 0.548$ & $0.177 \pm 0.019$ & $0.991 \pm 0.004$ \\
\basex{GradNorm}     & Laplacian & $0.293 \pm 0.009$ & $2.326 \pm 0.345$ & $0.162 \pm 0.010$ & $0.996 \pm 0.001$ \\
\basex{IMTL-G}       & Laplacian & $0.306 \pm 0.177$ & $167.8 \pm 179.4$ & $0.047 \pm 0.065$ & $0.989 \pm 0.016$ \\
\basex{ConFIG}       & Laplacian & $0.302 \pm 0.005$ & $2.798 \pm 0.050$ & $0.154 \pm 0.013$ & $0.994 \pm 0.003$ \\
\lantern             & Laplacian & $0.259 \pm 0.003$ & $1.287 \pm 0.036$ & $0.213 \pm 0.004$ & $0.983 \pm 0.002$ \\
\bottomrule
\end{tabular}
\end{table*}
Table~\ref{tab:prdc} reports the full PRDC breakdown behind the
mode-collapse finding of Section~\ref{sec:results-discussion}, resolving the
aggregate distances FPD, KPD, and CFD into separate fidelity and
diversity components. Precision and density measure fidelity; recall
and coverage measure diversity~\cite{kynkaanniemi2019improved,
naeem2020reliable}. Density is normalized so that agreement with the
reference yields values near one, and values well above one indicate
over-concentration in dense regions of the reference distribution
rather than higher fidelity. \base and \lantern keep these
balanced, with density near $1.3$ and recall near $0.21$, close to
CaloDREAM. The general-purpose multi-task methods instead raise
density and precision while lowering recall, the signature of mode
collapse: samples concentrate in high-density regions and stop
covering the full reference distribution. \basex{IMTL-G} is the most
severe. Under the Laplacian loss its density reaches $167.8$ while
recall falls to $0.047$, so its samples pile onto a small region and
miss most of the reference. \basex{ConFIG} and \basex{GradNorm} show
the same trend more mildly, with density above $2$ and recall below
$0.18$. The high precision of these methods is therefore a symptom of
collapse, not a sign of quality. \lantern shows no such collapse: it
keeps the highest recall among the trained models, tied with \base,
and a density close to that of the reference-quality prior
surrogates. Among the prior surrogates CaloDiT shows the same
collapse, with density $10.2$ and recall $0.060$, consistent with its
poor FPD.

\paragraph{Schedule ablation.}
Complementing the cross-method comparison above,
Table~\ref{tab:schedule-ablation-prdc} reports PRDC for the three
auxiliary-loss schedules of Section~\ref{sec:ablation}. The voxel loss
shows the collapse signature (high density, low recall) under S1 and S2
that the terminal gate of S3 removes, while the Laplacian loss stays
near density~$1.3$ and recall~$0.21$ across all three schedules.

\begin{table}[t]
\centering
\caption{Precision, density, recall, and coverage under the auxiliary
loss schedules of Table~\ref{tab:schedule-ablation}.}
\label{tab:schedule-ablation-prdc}
\small
\setlength{\tabcolsep}{2.5pt}
\begin{tabular}{llcccc}
\toprule
Auxiliary & Sched. & Precision & Density & Recall & Coverage \\
\midrule
\multirow{3}{*}{Voxel}
  & S1 & $0.520$ & $6.471$ & $0.081$ & $0.992$ \\
  & S2 & $0.586$ & $10.090$ & $0.060$ & $0.987$ \\
  & S3 & $0.256$ & $1.309$ & $0.216$ & $0.981$ \\
\addlinespace[4pt]
\multirow{3}{*}{Laplacian}
  & S1 & $0.266$ & $1.342$ & $0.210$ & $0.983$ \\
  & S2 & $0.256$ & $1.255$ & $0.217$ & $0.982$ \\
  & S3 & $0.259$ & $1.287$ & $0.213$ & $0.983$ \\
\bottomrule
\end{tabular}
\end{table}

\subsection{Results for Weighted Sum and Uncertainty Weighting}
\label{app:wsum_uw_results}
Table~\ref{tab:model_metrics_1} separates the three combination rules cleanly.
Scheduled uncertainty weighting attains the lowest FPD, KPD, and CFD in both
representations and drives the classifier AUCs toward $0.5$, the point at which
generated and reference showers become hard to separate. In voxel space it
reaches an FPD of $115.02$ and a KPD of $0.384$, more than an order of magnitude
below weighted-sum ($2854.24$, $27.127$) and close to two orders below pure
uncertainty weighting ($8372.79$, $85.592$). The classifier tests agree: the
scheduled model sits at AUC $0.617$ and $0.571$ on the high- and low-level
features, whereas the unscheduled variants stay almost perfectly separable above
$0.97$.

The Laplacian representation narrows this gap by improving the
uncertainty-weighted models rather than weighted-sum. Pure UW falls from an FPD
of $8372.79$ in voxel space to $237.95$, and its high-level AUC drops from
$0.998$ to $0.628$, while weighted-sum remains poor (FPD $1518.22$, AUC
$0.976$). Scheduled uncertainty weighting again gives the strongest
distributional scores (FPD $148.77$, KPD $0.541$, CFD $0.148$), and the two
uncertainty variants become comparable on the classifier tests.

Two observations follow. The scheduled gate is what makes uncertainty weighting
usable, since without it pure UW is the worst model in voxel space on every
column. CFD follows the same ordering as FPD and KPD but on a compressed scale,
and it is the only metric on which the scheduled model improves over the
baselines in both representations, which supports its role as a
correlation-sensitive score that FPD and KPD do not capture. The low classifier
AUCs for the scheduled model should be read alongside its lower PRDC precision in
Table~\ref{tab:model_metrics_prdc}: it trades per-sample sharpness for a
distribution close enough to Geant4 to defeat the classifier.
\begin{table*}[t]
\centering
\caption{Model evaluation metrics. \basex{WSum} denotes weighted-sum loss
balancing, \basex{pure UW} uncertainty weighting without scheduling, and
\basex{sched UW} scheduled uncertainty weighting. CFD is computed in voxel
space. AUC~(HL) and AUC~(LL) are the high- and low-level classifier two-sample
tests, where values closer to $0.5$ indicate samples less distinguishable from
data.}
\label{tab:model_metrics_1}
\begin{tabular}{lcccccc}
\toprule
Model & Loss & FPD ($\times 10^{3}$) $\downarrow$ & KPD ($\times 10^{3}$) $\downarrow$ & CFD $\downarrow$ & AUC (HL) & AUC (LL) \\
\midrule
\basex{WSum}      & Voxel     & $2854.24 \pm 14.31$ & $27.127 \pm 1.531$ & $0.273$ & $0.991$ & $0.979$ \\
\basex{pure UW}   & Voxel     & $8372.79 \pm 21.14$ & $85.592 \pm 2.834$ & $0.351$ & $0.998$ & $0.991$ \\
\basex{sched UW}  & Voxel     & $115.02 \pm 2.79$   & $0.384 \pm 0.167$  & $0.145$ & $0.617$ & $0.571$ \\
\midrule
\basex{WSum}      & Laplacian & $1518.22 \pm 11.29$ & $12.571 \pm 0.913$ & $0.219$ & $0.976$ & $0.955$ \\
\basex{pure UW}   & Laplacian & $237.95 \pm 4.70$   & $1.096 \pm 0.291$  & $0.181$ & $0.628$ & $0.639$ \\
\basex{sched UW}  & Laplacian & $148.77 \pm 3.35$   & $0.541 \pm 0.178$  & $0.148$ & $0.645$ & $0.583$ \\
\bottomrule
\end{tabular}
\end{table*}

\begin{table*}[t]
\centering
\caption{PRDC Evaluation Metrics}
\label{tab:model_metrics_prdc}
\begin{tabular}{llcccc}
\toprule
Model & Loss & Precision~$\uparrow$ & Recall~$\uparrow$ & Density~$\uparrow$ & Coverage~$\uparrow$ \\
\midrule
\basex{WSum}     & Voxel     & $0.933$ & $0.011$ & $317.0$ & $0.930$ \\
\basex{pure UW}  & Voxel     & $0.732$ & $0.218$ & $437.5$ & $0.745$ \\
\basex{sched UW} & Voxel     & $0.297$ & $0.163$ & $2.32$  & $0.997$ \\
\midrule
\basex{WSum}     & Laplacian & $0.865$ & $0.009$ & $123.9$ & $0.954$ \\
\basex{pure UW}  & Laplacian & $0.269$ & $0.159$ & $3.10$  & $0.990$ \\
\basex{sched UW} & Laplacian & $0.280$ & $0.179$ & $2.12$  & $0.995$ \\
\bottomrule
\end{tabular}
\end{table*}
\paragraph{PRDC analysis.}
The density values far above unity in Table~\ref{tab:model_metrics_prdc}, such as $317$ and $437$ for the voxel weighted-sum and pure uncertainty models, do not indicate stronger fidelity; they show that generated samples pile into the densest region of the real manifold, where heavily overlapping neighborhood balls inflate the per-sample count. The accompanying near-zero recall ($0.011$ for weighted-sum voxel) confirms mode concentration, since the samples are realistic yet enclose almost none of the real distribution. Pure uncertainty weighting raises both precision and density by driving mass into these high-density regions, but its coverage drops to $0.745$ because the sparse tail modes receive few generated neighbors. The scheduled variant instead restores a density near $2$ and coverage above $0.99$ at the cost of lower precision, a deliberate exchange of fidelity for broader mode coverage. This behavior is sharpest under the voxel representation: the Laplacian loss keeps the uncertainty models near a density of $2$, whereas weighted-sum still collapses.
\subsection{Gradient Diagnostic Definitions}
\label{app:diag-definitions}

This section defines the quantities in Table~\ref{tab:rq4:ratios}. All are
computed from per-epoch gradient statistics logged during training: the norms of
the denoising gradient $g_d$, the auxiliary gradient $g_a$, and the applied
blended step $g_{\mathrm{blend}}$, together with the cosines of
$g_{\mathrm{blend}}$ with each objective and the conflict angle between $g_d$ and
$g_a$.

\paragraph{Progress ratios.}
For a smooth loss $L$ and a step $\Delta\theta$, the first-order change in $L$ is
$\langle \nabla L, \Delta\theta \rangle$. We compare the applied step
$\Delta\theta_a = -\eta\, g_{\mathrm{blend}}$ against a pure step on one objective
$\Delta\theta_n = -\eta\, g_k$ by the ratio of their first-order terms and of
their squared sizes,
\begin{equation}
R_1^{k} = \frac{\|g_{\mathrm{blend}}\|}{\|g_k\|}\,
          \cos\!\big(g_{\mathrm{blend}}, g_k\big),
\qquad
R_2^{k} = \left(\frac{\|g_{\mathrm{blend}}\|}{\|g_k\|}\right)^{2},
\label{eq:app:ratios}
\end{equation}
with $k \in \{d, a\}$ for the denoising and auxiliary objectives. The first-order
ratio $R_1^{k}$ is the share of that objective's progress the blended step
retains. It folds in both the step size and the alignment, so it falls when the
step points away from $g_k$. The second-order ratio $R_2^{k}$ is the squared
step-size ratio alone. Balance is $R_1^{d}/R_1^{a}$; a value above one means the
step retains more of the denoising progress than of the auxiliary progress.

\paragraph{Cosines and conflict angle.}
$\cos_{d}$ and $\cos_{a}$ are the cosines of the blended step $g_{\mathrm{blend}}$
with $g_d$ and $g_a$. $\theta_{\mathrm{conf}}$ is the angle between $g_d$ and
$g_a$ before blending, in degrees.

\paragraph{Active window.}
The auxiliary weight follows a schedule that ramps from zero, holds, and decays
back toward zero. Outside the held region the auxiliary gradient is near zero by
design, and the ratios in Eq.~\ref{eq:app:ratios} diverge from the small
denominator. We therefore restrict all statistics to the active window, defined
per epoch as those epochs where the auxiliary weight exceeds a fixed fraction of
its peak value. We use a floor fraction of $0.01$, which includes the ramp.

\paragraph{Spike filter.}
Within the active window the auxiliary gradient norm can transiently collapse,
which inflates $R_1^{a}$ and $R_2^{a}$ through the small denominator. We drop
active epochs where the auxiliary gradient norm falls below one quarter of its
active-window median, and report ratios on the remaining epochs. The \%spike
column is the fraction of active epochs removed by this filter.

\paragraph{Aggregation.}
The logged statistics are heavy-tailed, so we summarize each run by the median
over its active window. We then report the mean and standard deviation of these
per-run medians across three seeds. The energy network is held fixed; the three
seeds vary the shape network.
\end{document}